\documentclass[11pt]{article}

\usepackage[final]{acl}
\usepackage{times}
\usepackage{latexsym}

\usepackage[T1]{fontenc}

\usepackage[utf8]{inputenc}

\usepackage{microtype}

\usepackage{inconsolata}

\usepackage{graphicx}
\usepackage{amsmath}
\usepackage{enumitem}
\usepackage{amssymb}
\usepackage{booktabs}
\usepackage{algorithm}
\usepackage{algpseudocode}
\usepackage{multirow} 
\usepackage{fvextra}
\usepackage{stfloats}

\title{Learning to Remember: End-to-End Training of Memory Agents for Long-Context Reasoning}

\author{
  Kehao Zhang$^{1,3}$\thanks{This work was done during Kehao Zhang's internship at Li Auto Inc.},
  Shangtong Gui$^{4}$,
  Sheng Yang$^{4}$,
  Wei Chen$^{4}$,
  \textbf{Yang Feng}$^{1,2,3}$\thanks{ Corresponding author: Yang Feng.}\thanks{The code is available at \url{https://github.com/ictnlp/unified-memory-agent.git}.} \\
  \textsuperscript{\rm 1}{Key Laboratory of Intelligent Information Processing,} \\
  Institute of Computing Technology, Chinese Academy of Sciences (ICT/CAS) \\
  \textsuperscript{\rm 2}{Key Laboratory of AI Safety, Chinese Academy of Sciences} \\
  \textsuperscript{\rm 3}{University of Chinese Academy of Sciences, Beijing, China} \\
  \textsuperscript{\rm 4}{Li Auto Inc.} \\
  {$\;\;$\texttt{\href{mailto:zhangkehao22z@ict.ac.cn}{zhangkehao22z@ict.ac.cn},\href{mailto:fengyang@ict.ac.cn}{fengyang@ict.ac.cn}}}
}

\begin{document}
\maketitle
\begin{abstract}
Long-context LLMs and Retrieval-Augmented Generation defer state tracking and evidence consolidation to query time, which is brittle when facts evolve and answers depend on latent states. We introduce Unified Memory Agent (UMA) for a one-to-many setting: query-agnostic external memory is constructed once from a stream and reused across multiple future QA sessions. A single policy maintains a structured Memory Bank through CRUD operations and answers using both the Memory Bank and raw context. Task-Stratified GRPO uses the mean reward of QA trajectories branching from each sampled memory state to supervise memory maintenance, while normalizing memory and per-question QA groups separately. We also introduce Ledger-QA, a diagnostic benchmark for long-horizon state tracking over accumulated updates. At the 16k budget, UMA-Generalist achieves the highest average score among compared methods across the test-time-learning and accurate-retrieval benchmarks and transfers to Ledger-QA without task-specific training; UMA-Specialist further improves long-horizon tracking after task adaptation. These results support learned proactive memory management for long-context reasoning.
\end{abstract}

\section{Introduction}
\label{sec:intro}

Modern Large Language Models (LLMs) increasingly support contexts of hundreds of thousands of tokens, with some reaching million-token windows~\citep{deepseekai2024deepseekv3technicalreport,anthropic2026claude47,openai2026gpt55,kavukcuoglu2026gemini35,qwen2026qwen37,shen2025qwenlong}. Yet long-context modeling faces challenges beyond window size: standard attention scales quadratically with sequence length~\citep{vaswani2017attention}, making very long inputs costly, while longer inputs can exhibit lost-in-the-middle effects and broader \textit{context rot} even when relevant information is present~\citep{liu2024lost,hong2025context}.

\begin{figure*}[t]
    \centering
    \includegraphics[width=0.82\textwidth]{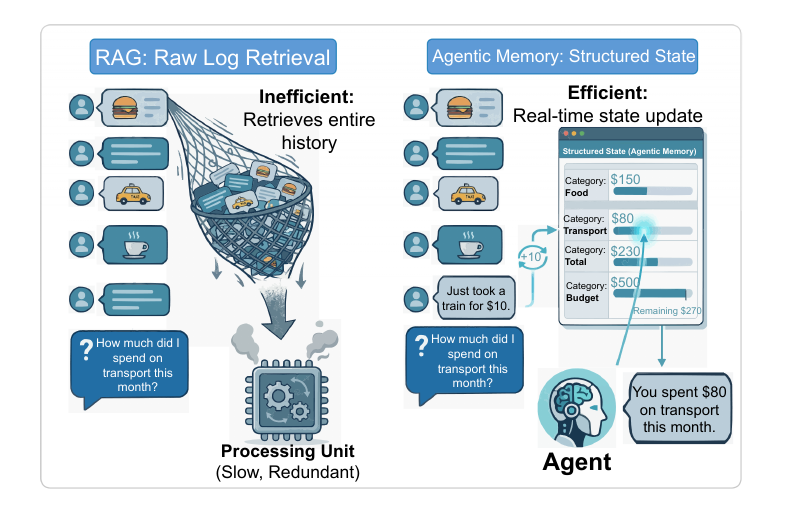}
    \caption{
        Expense-tracking example: RAG reprocesses logs per query, while agentic memory maintains structured state and answers from relevant fields.
    }
    \label{fig:ragvsam}
\end{figure*}

Retrieval-Augmented Generation (RAG) scales by retrieving relevant passages at query time, but shares the same \textbf{passive information processing} limitation with long-context approaches~\citep{xu2024retrieval, hu2025memory, zhang2025survey}: context is treated as static raw material and reasoning is deferred until a query arrives. As Figure~\ref{fig:ragvsam} shows, RAG reprocesses retrieved passages for every query; on state tracking, contradiction resolution, and evidence aggregation, this creates both accuracy degradation and computational redundancy.

\textbf{Memory-augmented agents} instead consolidate information as it arrives. Recent work explores RL-trained memory operations, downstream QA rewards, persistent structured stores, query-conditioned end-to-end memory-to-answer policies, and query-hidden memory construction~\citep{yu2025memagent,shi2025rememr1,yan2025memory,wang2025mem,yue2026mem,xu2025mem,yu2026agemem}. These advances expose a distinct optimization challenge: how to jointly train a reusable memory state against multiple future questions. In this \textit{one-to-many} setting, memory is constructed before the questions are revealed and subsequently shared by multiple independent QA sessions. The shared state must preserve broadly useful evidence rather than optimize for a single revealed objective, making supervision across questions central to training. Query-conditioned agents instead construct memory for a known question, whereas query-agnostic systems generally separate memory and QA optimization or associate each memory rollout with a single downstream task.

We propose \textbf{Unified Memory Agent (UMA)} to address this optimization problem. For each context, UMA samples memory-maintenance trajectories and branches each resulting memory state into multiple QA trajectories. Their mean utility provides a cross-stage, session-level learning signal for upstream memory maintenance, while \textbf{Task-Stratified GRPO} normalizes memory rollouts and each question's QA rollouts separately. This directly connects query-agnostic memory construction to the downstream utility of the shared state.

We also introduce \textbf{Ledger-QA}, a controlled diagnostic benchmark for continuous state tracking over accumulated updates. We evaluate UMA-Generalist on 13 established test-time-learning and accurate-retrieval benchmarks and on Ledger-QA without task-specific training, and evaluate UMA-Specialist on task-adapted Ledger-QA. At the 16k budget, UMA-Generalist achieves the highest aggregate TTL/AR score among compared methods and transfers to Ledger-QA, while UMA-Specialist further improves long-horizon tracking. Together, the results support proactive memory management as a complement to longer context windows and query-time retrieval.

\section{Method}

\begin{figure*}[t]
    \centering
    \includegraphics[width=1.0\linewidth]{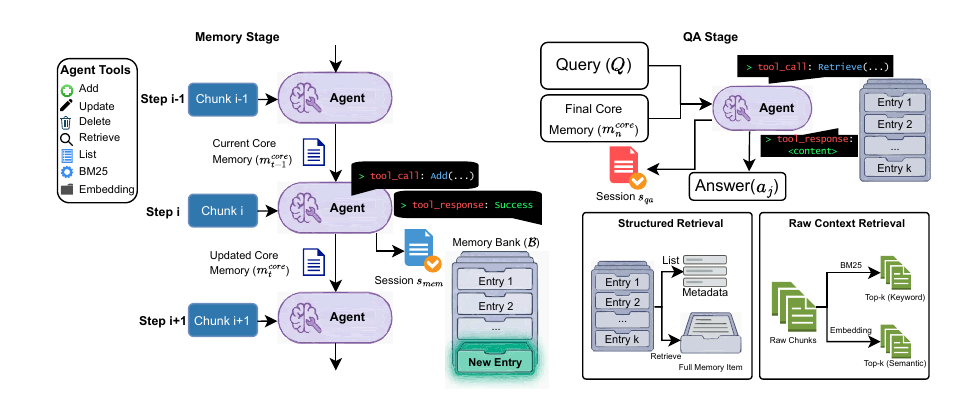}
    \caption{
        \textbf{Overview of UMA.} Phase I incrementally maintains a structured Memory Bank and core summary via CRUD over chunks. Phase II answers queries using both structured retrieval from the bank and raw-context retrieval.
    }
    \label{fig:mem_process}
\end{figure*}

\subsection{Problem Formulation as MDP}
\label{sec:formulation}

We formulate the long-context reasoning task as a Markov Decision Process (MDP). To handle sequences exceeding the context window, we decompose the input into a stream of chunks $C = \{c_1, \dots, c_N\}$. The MDP tuple $(\mathcal{S}, \mathcal{A}, \mathcal{P}, \mathcal{R})$ is defined as follows:

\paragraph{State Space ($\mathcal{S}$).} 
At step $t$, the state $s_t = (\mathcal{M}_t, x_t, h_t)$ consists of:
\begin{itemize}[leftmargin=*, itemsep=1pt, topsep=2pt]
    \item $\mathcal{M}_t = (m_t^{core}, \mathcal{B}_t)$: The dual-component memory, comprising a high-level summary ($m_t^{core}$) and a structured key-value memory bank ($\mathcal{B}_t$).
    \item $x_t$: The current input focus. To circumvent context window limits, this is dynamically switched between a local text chunk $c_k$ (during memory maintenance) and a specific query $q$ (during QA).
    \item $h_t$: The immediate interaction history, which buffers recent observations and is periodically cleared to manage context length.
\end{itemize}

\paragraph{Action Space ($\mathcal{A}$).}
We unify all operations into a single action space, categorized by the phase of interaction. Crucially, read-only tools are available in both phases to ensure the agent can verify beliefs or query the memory at any time. Detailed specifications for executable tools and terminal-action parsing conventions are provided in Appendix~\ref{app:tool_definitions}.

\begin{itemize}[leftmargin=*, itemsep=1pt, topsep=2pt]
    \item \textbf{Memory Maintenance Operations ($\mathcal{A}_{mem}$):} Used during the chunk processing phase to maintain the state.
    \begin{itemize}[leftmargin=1em, itemsep=0pt]
        \item \texttt{Add}, \texttt{Update}, \texttt{Delete}: Write operations that create, revise or rename, and remove entries in $\mathcal{B}_t$.
        \item \texttt{Retrieve}, \texttt{List}: Read-only operations for accessing entries and keys in $\mathcal{B}_t$.
        \item \texttt{UpdateCore}: A special terminal action that commits the current progress to $m_t^{core}$ and triggers the transition to the next chunk.
    \end{itemize}
    
    \item \textbf{QA Operations ($\mathcal{A}_{qa}$):} Used during the query phase to synthesize the answer.
    \begin{itemize}[leftmargin=1em, itemsep=0pt]
        \item \texttt{BM25}, \texttt{Embedding}: Ranked retrieval actions for lexical and semantic search.
        \item \texttt{Retrieve}, \texttt{List}: Read-only operations for accessing entries and keys in $\mathcal{B}_t$.
        \item \texttt{Answer}: The terminal action to generate the final response.
    \end{itemize}
\end{itemize}

\paragraph{Transition Dynamics ($\mathcal{P}$).}
The system evolves through two nested loops plus a final phase switch:
\begin{itemize}[leftmargin=*, itemsep=1pt, topsep=2pt]
    \item \textbf{Inner Loop (within one session/chunk):} The agent performs multiple tool calls before committing the chunk. Each step appends interaction traces (tool calls and observations) to $h_t$ and may update the memory bank $\mathcal{B}_t$, while the current input focus $x_t=c_k$ remains unchanged.
    \item \textbf{Outer Loop (across chunks):} A terminal \texttt{UpdateCore} action ends the current inner-loop episode, writes the consolidated update to $m_t^{core}$, advances $x_t$ from $c_k$ to $c_{k+1}$, and re-initializes $h_t$ for the next chunk.
    \item \textbf{Phase Transition:} After processing $c_N$, the system switches from maintenance to QA, where $x_t$ becomes a query $q$ sampled from $P(q|C)$.
\end{itemize}
This transition structure implies that the policy should both optimize short-horizon inner-loop edits on $\mathcal{B}_t$ and optimize outer-loop consolidation into $m_t^{core}$, so that the final memory state $\mathcal{M}_N$ supports future queries.

\subsection{Unified Memory Agent Architecture}
\label{sec:architecture}

Based on the MDP formulation, UMA employs a unified policy $\pi_\theta(a_t | s_t)$ to generate actions. The architecture is designed to handle the inputs explicitly as follows:

\paragraph{Input Representation.}
At any time step $t$, the input context provided to the LLM is a concatenation of four segments:
\begin{equation}
    \text{Input}_t = [I_{sys}, m_t^{core}, x_t, h_t]
\end{equation}
\begin{itemize}[leftmargin=*, itemsep=1pt, topsep=2pt]
    \item \textbf{System Instruction ($I_{sys}$):} Specifies the phase-specific operational guidelines—governing either memory maintenance or question answering—and defines the available toolset ($\mathcal{A}_{mem}$ or $\mathcal{A}_{qa}$).
    \item \textbf{Core Memory ($m_t^{core}$):} The evolving high-level summary that provides global context.
    \item \textbf{Current Focus ($x_t$):} The raw text of the current chunk $c_k$ (in Phase I) or the user query $q$ (in Phase II).
    \item \textbf{Interaction History ($h_t$):} The trajectory of recent tool calls and their execution results.
\end{itemize}

\noindent The Unified Memory Agent framework contains two phases, as illustrated in Figure~\ref{fig:mem_process}. 

\paragraph{Phase I: Sequential Memory Maintenance.}
For each chunk $c_k$, the agent initializes $h_t = \emptyset$. It iteratively samples actions
\begin{equation}
    a_t \sim \pi_\theta(\cdot | I_{sys}, m_t^{core}, c_k, h_t)
\end{equation}

The environment executes the tool and appends the result to $h_t$. This loop continues until the agent outputs \texttt{UpdateCore}, which compresses relevant insights into $m_t^{core}$ and advances the process to $c_{k+1}$.

\paragraph{Phase II: Hybrid Retrieval-Augmented QA.}
After processing the final chunk, the input focus $x_t$ switches to the query $q$. The agent utilizes the fully constructed memory $\mathcal{M}_N$ and an empty history $h_t$. It can invoke hybrid retrieval tools (exact-title access via \texttt{Retrieve} and ranked search via \texttt{BM25}/\texttt{Embedding}) to gather evidence. The process terminates when the agent generates the \texttt{Answer} action.

\subsection{Training: Task-Stratified GRPO}
\label{sec:training}
\begin{figure*}[t]
    \centering
    \includegraphics[
        width=0.95\linewidth,
    ]{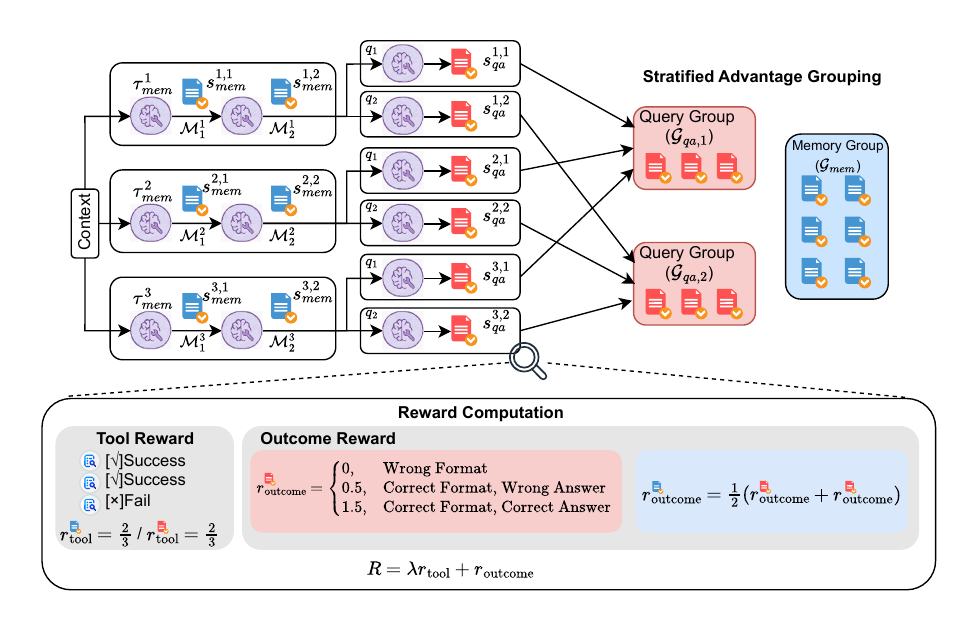}
    \caption{
        \textbf{Illustration of Task-Stratified GRPO.} 
        For a given input, we first sample multiple \emph{memory trajectories} (blue), each producing a final memory state; then we branch every state into \emph{QA sessions} (red) for a shared question set.
        \textbf{(Right)} Advantages are normalized within distinct groups: memory sessions share a memory group ($\mathcal{G}_{mem}$), while QA sessions are grouped by their specific question ($\mathcal{G}_{qa,j}$). 
        \textbf{(Bottom)} The reward combines immediate tool-execution feedback ($r_{tool}$) with outcome assessments ($r_{outcome}$). Each memory session receives the mean downstream QA outcome of its trajectory together with its own tool reward.
    }
    \label{fig:mem_rl}
\end{figure*}
We propose a \textbf{Task-Stratified Group Relative Policy Optimization} algorithm. By viewing GRPO as a Monte Carlo (MC) advantage estimator without a value network, we adapt it to handle the hierarchical structure of our task as illustrated in Figure~\ref{fig:mem_rl}.

\subsubsection{Nested Trajectory Sampling}
To connect query-agnostic memory construction with downstream QA utility, we employ a nested sampling strategy during training. For a given context $C$:
\begin{enumerate}[leftmargin=*, nosep]
    \item \textbf{Memory Rollout (trajectory level):} We sample $G$ trajectories $\{\tau^1, \dots, \tau^G\}$. Each $\tau^i$ comprises $N$ memory-tool-calling sessions $\{s_{mem}^{i,1}, \dots, s_{mem}^{i,N}\}$, one per context chunk, and yields a final state $\mathcal{M}_N^i$.
    \item \textbf{Shared Query Sampling:} Once per context, we sample $\mathcal{Q}_C=\{q_j\}_{j=1}^{M}$ and reuse it across all $\mathcal{M}_N^i$. Here, $M$ denotes the number of questions for the current training instance and varies across datasets.
    \item \textbf{QA Rollout (session level):} For each pair $(\mathcal{M}_N^i, q_j)$, we generate one multi-turn QA-tool-calling session $s_{qa}^{i,j}$ and compute its reward $R_{qa}^{i,j}$.
\end{enumerate}
Each update therefore contains $G$ memory trajectories and $GM$ QA sessions. We fix $G=16$; complete settings are in Appendix~\ref{app:train_config}.

\subsubsection{Reward Function}
Rewards are computed at the \emph{session} level, with different outcome assignments for QA and memory sessions. Accordingly, actions within the same tool-calling session share the same outcome-derived advantage.

For each QA session $s_{qa}^{i,j}$, the reward is
\begin{equation}
\begin{aligned}
    R_{qa}^{i,j} &= \lambda r_{tool}^{qa,i,j} + r_{outcome}^{i,j}, \\
    r_{outcome}^{i,j} &= r_{outcome}(a_{ans}^{i,j}, y_j)
\end{aligned}
\end{equation}
where $r_{tool}^{qa,i,j}$ is the tool success signal of this QA session.

For each memory session $s_{mem}^{i,k}$ in trajectory $\tau^i$, we keep an independent tool reward and use the averaged QA outcome under the same trajectory as its outcome signal:
\begin{equation}
\begin{aligned}
    \bar{r}_{outcome}^{i} &= \frac{1}{M}\sum_{j=1}^{M} r_{outcome}^{i,j}, \\
    R_{mem}^{i,k} &= \lambda r_{tool}^{mem,i,k} + \bar{r}_{outcome}^{i}
\end{aligned}
\end{equation}
where $r_{tool}^{mem,i,k}$ is computed from the tool-call validity of memory session $s_{mem}^{i,k}$ (e.g., valid-call ratio $N_{valid}/N_{total}$ for that session).
Here, $\lambda$ controls the trade-off between tool-use quality and answer outcome. We use the fixed setting $\lambda=0.5$ across tasks.

\subsubsection{Task-Stratified Advantage Estimation}
Given the session-level rewards defined above, we compute GRPO advantages via group-wise normalization. Our extension uses two stratified groups for memory and QA, respectively.

\paragraph{Advantage for Memory Sessions ($A_{mem}$).}
Each memory-session reward is normalized against all memory sessions sampled in the current update:
\begin{equation}
\begin{aligned}
    A_{mem}^{i,k} &= \frac{R_{mem}^{i,k} - \bar{R}_{mem}}{\sigma(R_{mem}^{\cdot,\cdot}) + \epsilon_{std}}, \\
    \bar{R}_{mem} &= \frac{1}{GN}\sum_{u=1}^{G}\sum_{v=1}^{N}R_{mem}^{u,v}
\end{aligned}
\end{equation}
The resulting advantage is assigned to the session, and all actions within that tool-calling session share it.

\paragraph{Advantage for QA Sessions ($A_{qa}$).}
For a specific question $q_j$, the difficulty is intrinsic. Therefore, we normalize each QA-session reward only against other sessions answering the \textit{same} question $q_j$ (but with different constructed memory banks).
\begin{equation}
    A_{qa}^{i,j} = \frac{R_{qa}^{i,j} - \bar{R}_{qa,j}}{\sigma(R_{qa}^{\cdot,j}) + \epsilon_{std}},
    \quad
    \bar{R}_{qa,j}=\frac{1}{G}\sum_{u=1}^{G}R_{qa}^{u,j}
\end{equation}
Here, $\epsilon_{std}=10^{-6}$ prevents division-by-zero errors. Normalizing within each question group removes cross-question difficulty variation, so advantages reflect relative performance among rollouts for the same question.

Finally, the policy $\pi_\theta$ is updated by maximizing a unified GRPO objective over all sampled sessions:
\begin{equation}
\mathcal{J}(\theta)=\frac{1}{GN+GM}\big(\mathcal{S}_{mem}+\mathcal{S}_{qa}\big)
\end{equation}
\begin{equation}
\begin{aligned}
\mathcal{S}_{mem} &= \sum_{i=1}^{G}\sum_{k=1}^{N}
\frac{1}{|s_{mem}^{i,k}|}\sum_{t\in s_{mem}^{i,k}} \ell(\rho_t, A_{mem}^{i,k}), \\
\mathcal{S}_{qa} &= \sum_{i=1}^{G}\sum_{j=1}^{M}
\frac{1}{|s_{qa}^{i,j}|}\sum_{t\in s_{qa}^{i,j}} \ell(\rho_t, A_{qa}^{i,j})
\end{aligned}
\end{equation}
\begin{equation}
\begin{aligned}
\ell(\rho_t, A) ={}& \min\!\big(\rho_t A,\, \text{clip}(\rho_t,1-\epsilon,1+\epsilon)A\big) \\
&- \beta D_{KL,t}
\end{aligned}
\end{equation}
Here, $\rho_t = \frac{\pi_\theta(a_t\mid s_t)}{\pi_{\theta_{\mathrm{old}}}(a_t\mid s_t)}$ denotes the importance sampling ratio between the current policy $\pi_\theta$ and the rollout policy $\pi_{\theta_{\mathrm{old}}}$ used for trajectory collection. In contrast, $D_{KL,t}=D_{KL}\big(\pi_\theta(\cdot\mid s_t)\,\|\,\pi_{\theta_{\mathrm{ref}}}(\cdot\mid s_t)\big)$ penalizes divergence from the frozen reference policy $\pi_{\theta_{\mathrm{ref}}}$.
\section{The Ledger-QA Benchmark}
\label{sec:ledger_qa}

We introduce \textbf{Ledger-QA}, a controllable and extensible synthetic benchmark for long-horizon state tracking. Existing long-term memory benchmarks such as LongMemEval~\citep{wulongmemeval} evaluate retrieval and reasoning over extended interaction histories, including temporal reasoning and knowledge updates. Ledger-QA targets a complementary challenge: continuously maintaining an accumulated latent state whose answer need not appear explicitly in any single span. Rather than claiming full realism, it uses simple, transparent generation rules to isolate this capability while remaining challenging for modern long-context agents.

\subsection{Dataset Construction}

The construction pipeline contains three stages:

\paragraph{Timeline \& Intent Sampling.}
We sample $H$ dates from \texttt{2023-01-01} to \texttt{2024-12-31} and select consumption-scene intents for each session, forming a chronological stream of date-stamped updates.

\paragraph{Joint Dialogue \& Transaction Synthesis.}
Given each date and its scene intents, a strong LLM (e.g., Gemini 3 Pro~\citep{pichai2025gemini3}) jointly generates natural user-assistant dialogue and aligned structured \texttt{Transaction} records, including context-appropriate amounts. Amounts are not sampled beforehand but determined during this joint synthesis, and the two outputs are returned together to preserve their semantic alignment.

\paragraph{Ground-Truth Query Generation.}
From the resulting structured ledger, we programmatically generate questions and exact answers across 8 categories that require multi-step aggregation over accumulated updates rather than local span matching.

A concrete construction example and full generation details are provided in Appendix~\ref{app:ledger_construction}.

\subsection{Why Ledger-QA is Challenging}
Ledger-QA fundamentally differs from standard RAG benchmarks by demanding \textbf{long-horizon state aggregation} rather than local information retrieval. Answers (e.g., "Total spending") are latent values derived from summing scattered transactions ($v_1 + \dots + v_T$) across the timeline. This requires continuous belief-state maintenance: local extraction errors can propagate to final answers, and simple retrieval or approximate summarization is often insufficient. Ledger-QA should therefore be viewed as a controlled diagnostic benchmark rather than a complete proxy for real-world memory use: its simple, transparent rules isolate the prerequisite of consolidating sequential updates into a persistent state. Strong performance does not by itself guarantee robustness in noisier real-world scenarios, but poor performance reveals a weakness in this foundational capability.
\section{Experiment}

\subsection{Experimental Setup}

\subsubsection{Evaluation Benchmarks}
\label{sec:benchmarks}

We evaluate UMA along three primary axes of memory behavior: \textbf{Dynamic State Tracking} (our Ledger-QA in Section~\ref{sec:ledger_qa}), \textbf{Test-Time Learning (TTL)} (\textit{TREC-Coarse}, \textit{TREC-Fine}, \textit{NLU}, \textit{Clinic}, \textit{Banking77}, \textit{PubMed-RCT}), and \textbf{Accurate Retrieval (AR)} (\textit{HotpotQA}, \textit{LoCoMo}, \textit{LongMemEval}, \textit{MSC}, \textit{PerLTQA}, \textit{SQuAD}, \textit{ConvoMem})~\citep{hu2025evaluating,yang2018hotpotqa,maharana2024evaluating,wulongmemeval,xu2022beyond,du2024perltqa,rajpurkar2016squad,pakhomov2025convomem}.

\paragraph{Task focus.}
Ledger-QA tests persistent state maintenance under continual updates; TTL tests storage and application of task rules from in-context supervision; AR evaluates fine-grained factual recovery from long documents or conversations. Together, they cover distinct memory capabilities rather than repeatedly testing a single aspect.

\paragraph{Evaluation metric.}
We use an LLM-as-a-Judge protocol with \texttt{DeepSeek-V3.2} and additionally report deterministic lexical metrics using a shared final-answer extraction and normalization pipeline. Evaluation details, prompt templates, and dataset taxonomies are provided in Appendices~\ref{app:detailed_metrics}, \ref{app:eval_prompts}, and~\ref{app:data_details}.

\subsubsection{Baselines}
\label{sec:baselines}
We compare UMA against a diverse set of baselines. First, we evaluate standard approaches: \textbf{Concat}, which directly concatenates the context up to the context budget to test the backbone's native capacity, and \textbf{RAG}, which retrieves the top-20 chunks using a hybrid strategy combining dense embeddings and sparse BM25 via Reciprocal Rank Fusion (RRF; \citealp{cormack2009reciprocal}). Second, we include recurrent summarization models: \textbf{MemAgent}~\citep{yu2025memagent} iteratively updates a running memory state based on the query, while \textbf{MemAgent-woq} performs query-agnostic compression. Finally, we compare against advanced memory frameworks: \textbf{Mem1}~\citep{zhou2026mem1} interleaves chunk reading with dynamic retrieval to maintain an evolving internal state, \textbf{A-MEM}~\citep{xu2025mem} is an additional external-memory baseline with explicit memory operations, \textbf{Mem-$\alpha$}~\citep{wang2025mem} utilizes an RL-optimized policy to manage a composite memory system (core, episodic, and semantic), and \textbf{Mem-T}~\citep{yue2026mem} is a hierarchical memory agent trained in two stages: optimizing retrieval via tree-guided RL and then refining its construction policy through hindsight-based distillation. Implementation details are provided in Appendix~\ref{app:base_details}.

\subsubsection{Implementation Details}
We train two checkpoints with the same architecture and optimization. \textbf{UMA-Generalist} is trained for two epochs on modified HotpotQA and a filtered Mem-$\alpha$ split containing SQuAD, PerLTQA, NLU, TREC-Coarse, PubMed-RCT, and BookSum, with HotpotQA and LongMemEval removed. It is used for TTL/AR and evaluated on Ledger-QA without task-specific training. \textbf{UMA-Specialist} is trained for six epochs on the independently generated Ledger-QA set. Dataset and filtering details are in Appendix~\ref{app:train_data_details}. Dense retrieval uses \texttt{sentence-transformers/all-MiniLM-L6-v2} served via \texttt{infinity}~\citep{feil_2023_11630143}. UMA builds on customized \texttt{veRL}~\citep{sheng2025hybridflow}, initializes from \texttt{Qwen/Qwen3-4B-Instruct-2507}\footnote{We use the Instruct checkpoint because Qwen3-4B/8B thinking variants sometimes generate overlong traces that fill the context window, making rollout-based RL costly.}, and trains Task-Stratified GRPO on 32 NVIDIA H200 GPUs. Hyperparameters and default prompts are provided in Appendices~\ref{app:train_config} and~\ref{app:agent_prompts}.

\subsection{Main Results}
\label{sec:main_results}

\begin{table*}[t]
    \centering
    \resizebox{\textwidth}{!}{
        \small
        \setlength{\tabcolsep}{3pt}
        \renewcommand{\arraystretch}{1.15}
        \begin{tabular}{l|cccccc|ccccccc|c}
        \toprule
        \multirow{2}{*}{\textbf{Method}} & \multicolumn{6}{c|}{\textbf{Test-Time Learning (TTL)}} & \multicolumn{7}{c|}{\textbf{Accurate Retrieval (AR)}} & \\
        & \textbf{Bank77} & \textbf{Clinic} & \textbf{NLU} & \textbf{Pub} & \textbf{T-C} & \textbf{T-F} & \textbf{Convo} & \textbf{Hotpot} & \textbf{LoCo} & \textbf{LME} & \textbf{MSC} & \textbf{PerL} & \textbf{SQuAD} & \textbf{Avg.} \\
        \midrule
        \multicolumn{15}{l}{\textit{Baselines}} \\
        Concat & 72.00 & 79.00 & 71.00 & 57.80 & 71.00 & 18.00 & 2.46 & 63.28 & 30.01 & 31.60 & 58.40 & 36.00 & 63.38 & 50.30 \\
        RAG & 80.00 & 70.00 & 65.00 & 51.20 & 9.00 & 14.00 & 51.12 & 78.12 & 54.03 & \textbf{59.60} & 54.60 & \textbf{83.75} & 25.35 & 53.52 \\
        MemAgent & 25.00 & 13.00 & 22.00 & 49.90 & 61.00 & 37.00 & 59.60 & 51.56 & \textbf{60.37} & 55.40 & 50.20 & 65.25 & 77.02 & 48.25 \\
        MemAgent-woq & 27.00 & 41.00 & 34.00 & 44.40 & 70.00 & 44.00 & 1.56 & 11.72 & 25.43 & 7.60 & 29.40 & 23.50 & 25.01 & 29.59 \\
        Mem1 & 0.00 & 6.00 & 0.00 & 9.00 & 7.00 & 0.00 & 1.34 & 5.47 & 21.60 & 6.40 & 8.40 & 7.25 & 15.09 & 6.73 \\
        A-MEM & 83.00 & 81.00 & 72.00 & 53.70 & 36.00 & 16.00 & 52.46 & 67.97 & 45.87 & 31.80 & 62.80 & 73.00 & \textbf{88.12} & 58.75 \\
        Mem-$\alpha$ & 83.00 & 77.00 & 68.00 & 56.50 & 78.00 & 62.00 & 7.14 & \textbf{79.69} & 40.85 & 53.80 & 56.80 & 74.50 & 81.92 & 63.02 \\
        Mem-T & 67.00 & 52.00 & 58.00 & 46.80 & 15.00 & 7.00 & 51.12 & 66.41 & 41.04 & 52.40 & 65.80 & 56.75 & 67.03 & 49.72 \\
        \midrule
        \multicolumn{15}{l}{\textit{Ablations}} \\
        UMA (w/o RL \& Phase I) & 26.00 & 27.00 & 23.00 & 39.30 & 13.00 & 12.00 & 79.02 & 64.06 & 50.40 & 38.60 & 36.20 & 57.25 & 82.33 & 42.17 \\
        UMA (w/o Phase I) & 84.00 & 80.00 & 75.00 & 47.40 & 81.00 & 28.00 & 86.16 & 69.53 & 44.21 & 48.80 & 27.40 & 59.00 & 83.47 & 62.61 \\
        UMA (w/o RL) & 77.00 & 72.00 & 61.00 & 38.20 & 69.00 & 20.00 & 84.82 & 64.84 & 44.26 & 48.60 & 63.80 & 51.50 & 81.19 & 59.71 \\
        UMA (w/o Raw-Context) & 69.00 & 71.00 & 59.00 & 36.30 & 71.00 & 44.00 & 73.12 & 60.15 & 40.59 & 35.80 & 62.20 & 52.00 & 76.52 & 57.74 \\
        UMA (Global Group) & 69.00 & 77.00 & \textbf{84.00} & \textbf{72.50} & 88.00 & 53.00 & 81.47 & 60.94 & 56.55 & 37.60 & \textbf{74.40} & 73.00 & 81.98 & 69.96 \\
        UMA (Two-Stage) & 86.00 & 84.00 & 70.00 & 44.60 & 77.00 & 43.00 & 84.82 & 67.19 & 42.75 & 49.20 & 65.60 & 60.75 & 65.59 & 64.65 \\
        \midrule
        \textbf{UMA-Generalist (16k)} & \textbf{87.00} & \textbf{91.00} & 82.00 & 67.20 & \textbf{94.00} & \textbf{75.00} & \textbf{87.05} & \underline{76.56} & 46.58 & \underline{51.60} & 66.20 & \underline{74.50} & \underline{87.53} & \textbf{75.86} \\
        \midrule
        \multicolumn{15}{l}{\textit{Session-budget scaling}} \\
        UMA-Generalist (8k) & 79.00 & 88.00 & 75.00 & 65.60 & 95.00 & 70.00 & 80.80 & 64.06 & 44.91 & 44.80 & 64.60 & 71.25 & 85.81 & 71.45 \\
        UMA-Generalist (32k) & \underline{90.00} & \underline{92.00} & \underline{84.00} & \underline{68.60} & \underline{97.00} & \underline{83.00} & \underline{89.06} & 74.22 & \underline{47.48} & 49.20 & \underline{67.00} & 69.00 & 87.32 & \underline{76.76} \\
        \bottomrule
        \end{tabular}
    }
    \caption{
        DeepSeek-V3.2 LLM-as-a-Judge scores for UMA-Generalist, its ablations, and session-budget variants. Bold marks the best result among methods using the 16k budget; underlining marks the best result among the UMA-Generalist 8k, 16k, and 32k variants. \textbf{Pub}: PubMed-RCT; \textbf{Convo}: ConvoMem; \textbf{LoCo}: LoCoMo; \textbf{LME}: LongMemEval; \textbf{PerL}: PerLTQA; \textbf{T-C/T-F}: TREC-Coarse/Fine. See Section~\ref{sec:ablation} and Appendix~\ref{app:detailed_metrics} for details.
    }
    \label{tab:main_results}
\end{table*}

Table~\ref{tab:main_results} presents UMA-Generalist and the baselines across 13 standard benchmarks (6 TTL and 7 AR tasks).

\paragraph{Results on Test-Time Learning (TTL).} 
UMA-Generalist averages 82.70 across the six TTL tasks, the highest six-task average at the 16k budget, and leads four tasks. Its performance on Banking77 and Clinic, which are unseen during training, also indicates transfer to new classification domains.
Performance is also stable across classification formats, suggesting that the learned maintenance policy is not tied to a single label inventory.

\paragraph{Results on Accurate Retrieval (AR).} 
UMA-Generalist obtains the highest seven-task average of 70.00 and ranks among the top three methods on five tasks. It leads ConvoMem and ranks among the top three methods on HotpotQA, MSC, PerLTQA, and SQuAD. The aggregate gain reflects consistently strong performance across tasks. Efficiency and scalability analysis is provided in Appendix~\ref{app:efficiency_scalability}.
These benchmarks reward retaining evidence that remains useful across heterogeneous questions rather than compressing context toward one known target.

MemAgent-woq and Mem1 are notably weaker on several TTL tasks, where memory is constructed before the target classification query is revealed. This protocol tests whether compact states learned for objective-conditioned use transfer to query-hidden retention.

\subsection{Results on Ledger-QA}
\label{sec:res_ledger}

\begin{table}[t]
    \centering
    \small
    \setlength{\tabcolsep}{2.5pt}
    \renewcommand{\arraystretch}{1.1}
    \resizebox{\columnwidth}{!}{
    \begin{tabular}{lcccccc}
    \toprule
    \textbf{Model} & \textbf{LQA Train} & \textbf{Judge} & \textbf{EM} & \textbf{F1} & \textbf{R-L} & \textbf{Sub-EM} \\
    \midrule
    UMA (w/o RL) & No & 30.45 & 19.09 & 20.15 & 20.15 & 42.59 \\
    MemAgent & No & 40.52 & 19.28 & 23.63 & 23.64 & 42.50 \\
    UMA-Generalist (16k) & No & 38.18 & 34.39 & 36.24 & 36.24 & 43.42 \\
    UMA-Specialist (16k) & Yes & \textbf{53.02} & \textbf{49.39} & \textbf{51.62} & \textbf{51.62} & \textbf{55.65} \\
    \bottomrule
    \end{tabular}
    }
    \caption{Average Ledger-QA results (\%). \textbf{Judge} uses DeepSeek-V3.2; \textbf{LQA Train} indicates whether Ledger-QA training data is used.}
    \label{tab:ledger_checkpoint_summary}
\end{table}

\begin{figure}[t]
    \centering
    \includegraphics[width=1.0\linewidth]{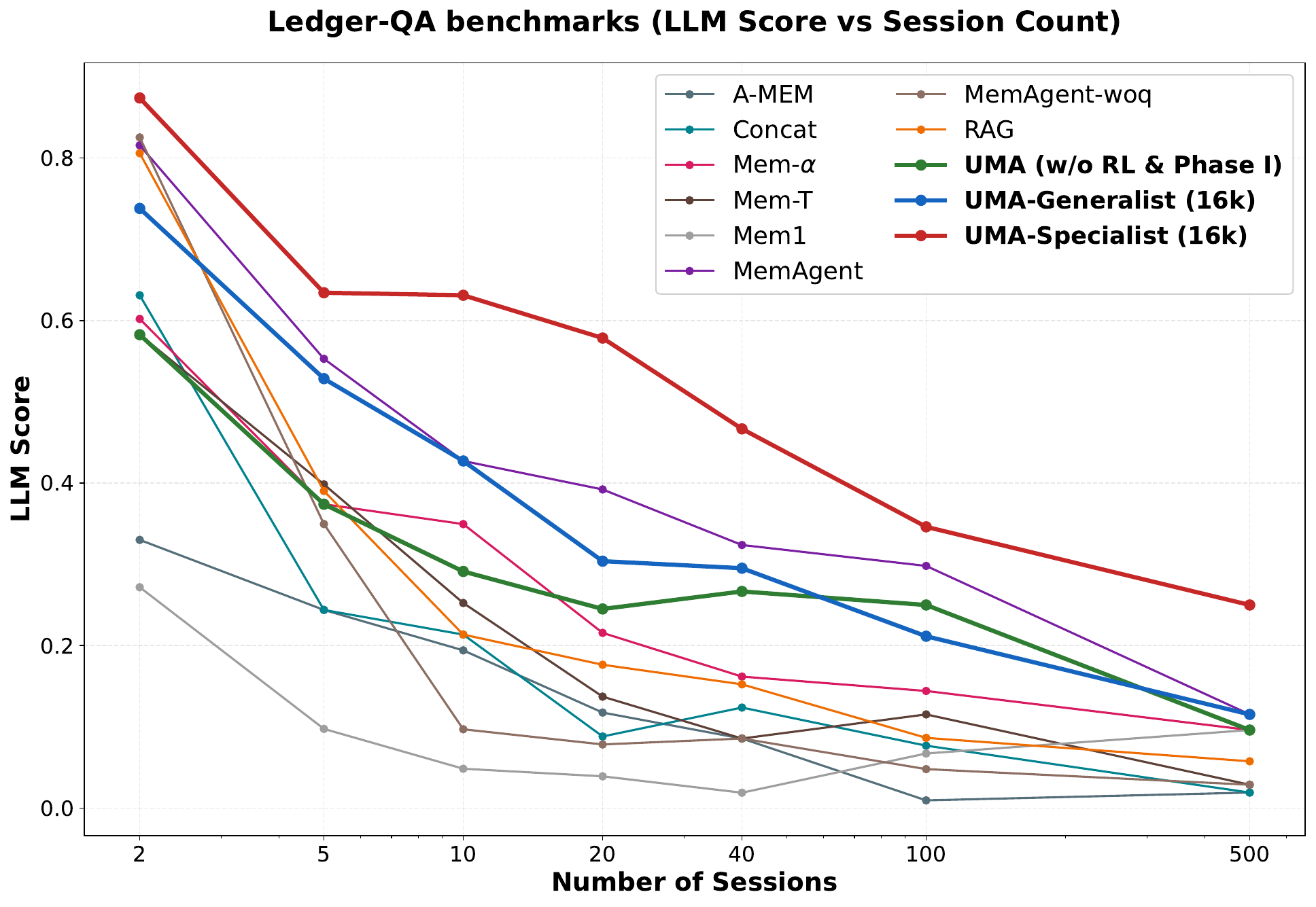}
    \caption{
        \textbf{Performance comparison on Ledger-QA across varying session horizons.} 
        The x-axis represents the dialogue-session horizon $H$, and the y-axis denotes LLM-as-a-Judge score. 
        The figure includes the evaluated baselines and multiple UMA variants; full results are in Table~\ref{tab:ledger_detailed}.
    }
    \label{fig:ledger_results}
\end{figure}

We evaluate Ledger-QA across session horizons $H \in \{2, 5, 10, 20, 30, 40, 50, 100, 500\}$. Figure~\ref{fig:ledger_results} compares the baselines and multiple UMA variants at representative horizons. Table~\ref{tab:ledger_checkpoint_summary} reports averages across all horizons and separates transfer from task adaptation; per-horizon Judge and EM results are provided in Appendix~\ref{app:ledger_results}.

Without Ledger-QA training, UMA-Generalist reaches 38.18 LLM-as-a-Judge and 34.39 EM, compared with 30.45 and 19.09 for UMA (w/o RL). MemAgent records a higher Judge score of 40.52 under its query-conditioned protocol, which constructs a separate state after seeing each question. UMA-Generalist reuses one query-hidden memory state across all questions and achieves higher EM, F1, and ROUGE-L. Task adaptation further raises UMA-Specialist to 53.02 Judge and 49.39 EM. The lexical metrics provide a complementary view because they require normalized answer overlap, whereas the Judge can accept semantically correct answers with different surface forms.

\paragraph{Vulnerability of Baselines.}
RAG and MemAgent-woq remain competitive at $H=2$ but fall to 5.77 and 2.88 at $H=500$; even query-conditioned MemAgent drops to 11.54. These declines show that retrieval and simple summarization struggle to aggregate scattered updates over long histories.

\paragraph{Long-Horizon Performance of UMA-Specialist.}
UMA-Specialist remains more stable, especially for $H \ge 20$. At 500 sessions, it scores 25.00, matching UMA (w/o Raw-Context) and exceeding the strongest non-UMA baseline, MemAgent, at 11.54. This margin supports explicit memory operations for accumulated state at ultra-long horizons.

\paragraph{Arithmetic Control.}
We further run a 100,000-example arithmetic-only control (Appendix~\ref{app:data_details}) where the model sums five random two-decimal numbers below 1000. Accuracy changes little after UMA-Specialist training (base: 25.41\%; UMA-Specialist: 25.14\%), suggesting Ledger-QA gains come from memory maintenance and state access rather than improved standalone arithmetic.
We provide a manual failure breakdown in Appendix~\ref{app:error_analysis}.

\subsection{Ablation Studies}
\label{sec:ablation}

We ablate architecture and training choices in the lower block of Table~\ref{tab:main_results}.

\paragraph{Ablation Settings.}
\textbf{w/o Phase I} removes sequential memory maintenance and relies on QA-time retrieval only. \textbf{w/o RL} keeps the full architecture but removes RL optimization. \textbf{w/o RL \& Phase I} removes both mechanisms. \textbf{w/o Raw-Context} disables \texttt{BM25}/\texttt{Embedding} and only allows Memory Bank retrieval at inference. \textbf{Global Group} normalizes all memory and QA sessions in a single global group without task stratification. \textbf{Two-Stage} separately trains memory and QA. \textbf{8k/32k Window} changes the default 16k session budget covering the initial prompt and later interaction tokens.

\paragraph{Impact of Memory Maintenance (Phase I).}
Explicit maintenance improves both untrained and RL models. Before RL, enabling write/update operations outperforms the retrieval-only variant, showing that structured state maintenance already benefits retention. After RL, full UMA exceeds \textbf{w/o Phase I} by 13.3 points on average, confirming that these operations contribute beyond QA-time retrieval.

\paragraph{Impact of RL Training.}
RL improves both architectures. It strengthens retrieval-only tool use and raises full UMA by 16.2 points on average, with gains from better memory writing, updating, and access.
The improvements span both retention-heavy TTL tasks and retrieval-heavy AR tasks instead of concentrating in one benchmark family.

\paragraph{Impact of Stratified Advantage Grouping.}
The \textbf{Global Group} variant, which normalizes all memory and QA sessions in a single global group, is 5.9 points lower on average. Separate normalization better matches binary QA outcomes and aggregated memory utility.

\paragraph{Two-Stage vs. Unified End-to-End Training.}
Two-stage training, which optimizes memory maintenance and QA separately, is 11.2 points lower on average. Joint optimization better links memory construction to downstream QA utility.

\paragraph{Effect of Session Context Budget.}
Moving from 8k to 16k clearly improves performance, whereas 32k provides only modest additional gains at higher compute and memory cost. We therefore retain 16k as the default trade-off.

\paragraph{Ablation: w/o Raw-Context Retrieval.}
Disabling raw-context retrieval produces a substantial drop on retrieval-heavy TTL/AR tasks and a small decrease on Ledger-QA. Across 952 Ledger-QA QA traces, UMA-Specialist accesses structured memory in every trace: 878 (92.2\%) use only Memory Bank tools, 74 (7.8\%) additionally invoke raw-context retrieval, and no trace uses raw-context retrieval alone. This pattern indicates that structured memory is the default access path on Ledger-QA, with raw retrieval used as an occasional supplement.

\section{Related Work}
\label{sec:related_work}

\noindent \textbf{Internal Memory and Context Compression.} 
Several approaches extend or compress context through internal representations, including parameter-based storage~\citep{wangself}, generative or learned memory tokens and slots~\citep{zhang2025memgen,burtsev2020memory,ge2024context}, and KV-based compression~\citep{qian2025memorag,li2024snapkv}. Such internal-memory approaches generally require model-level access or modification, whereas external memory systems can remain model-agnostic.

\noindent \textbf{Prompt-based External Memory.} 
Prompt-based systems such as MemGPT~\citep{packer2023memgpt}, Mem0~\citep{chhikara2025mem0}, and SCM~\citep{wang2026scm} maintain external stores through OS-like hierarchies, extracted memories, or graphs. Although black-box compatible, they depend on instruction following, making smaller models unreliable for complex memory updates and retrieval in dynamic settings~\citep{wang2025mem}.

\noindent \textbf{Trainable Memory Agents.} 
Recent work replaces hand-designed memory routines with learned policies. MEM1 interleaves retrieval with reasoning~\citep{zhou2026mem1}. MemAgent~\citep{yu2025memagent} and ReMemR1~\citep{shi2025rememr1} optimize question-conditioned memory-to-answer trajectories, with the target question guiding memory construction. A-MEM~\citep{xu2025mem} organizes an evolving structured store without policy training. Among query-agnostic learned-memory systems, Memory-R1~\citep{yan2025memory} trains specialized Memory Manager and Answer Agent policies; Mem-$\alpha$~\citep{wang2025mem} evaluates one shared memory against multiple questions while optimizing the memory writer through a fixed RAG-based QA pipeline; Mem-T~\citep{yue2026mem} combines hierarchical memory construction and multi-step retrieval with tree-guided rewards; and AgeMem~\citep{yu2026agemem} jointly learns long- and short-term memory actions with task execution in a three-stage trajectory.

Appendix~\ref{app:memory_agent_comparison} summarizes these formulations. UMA jointly optimizes one policy over query-agnostic memory construction and multiple downstream QA trajectories, with separate normalization for memory and per-question QA groups.

\section{Conclusion}
\label{sec:conclusion}

We propose Unified Memory Agent, which jointly optimizes memory acquisition and utilization via reinforcement learning. Task-Stratified GRPO aligns the objectives of long-horizon memory maintenance and downstream question answering. We also introduce \textbf{Ledger-QA}, a dataset requiring persistent updates to an evolving ledger. At the 16k budget, UMA-Generalist achieves the highest aggregate test-time-learning and accurate-retrieval score among compared methods, and UMA-Specialist improves long-horizon state tracking on Ledger-QA. Ablation studies show the contributions of policy training and explicit memory updates. Together, these results distinguish transfer from task-specific adaptation and support proactive memory management alongside longer contexts and query-time retrieval.

\section*{Limitations}
\label{sec:limitations}

Several limitations remain. First, we study single-agent settings with a fixed tool interface; collaborative agents and evolving APIs are left for future work. Second, Ledger-QA is a controlled synthetic diagnostic with transparent generation rules, and broader validation on real-world proactive state-maintenance tasks is still needed. Third, the latency and scaling measurements reflect the reported hardware and workloads. Finally, downstream outcomes supervise memory maintenance at the session level, so actions within one tool-calling session share an advantage; identifying the contribution of individual CRUD operations would require finer-grained or counterfactual attribution. Persistent memory also introduces privacy and data-retention risks, discussed below.

\section*{Ethical Considerations}

Long-term memory agents may retain sensitive personal information across sessions. Deployments should obtain informed consent, minimize stored content, apply retention limits and access controls, and provide users with mechanisms to inspect, correct, and delete memory. Sensitive information should be excluded by default unless its storage is necessary and explicitly authorized. Deletion should cover both memory entries and derived retrieval artifacts, including indexes and caches. Ledger-QA is fully synthetic and contains no real personal or financial records.

\section*{Acknowledgments}

This work was supported by the National Natural Science Foundation of China (General Program, Grant No. 62376260) and the Beijing Natural Science Foundation (Grant No. L257006).

We used ChatGPT (OpenAI) for language polishing and manuscript consistency checks. It was also used as an auxiliary tool during reference review; all references were manually verified by the authors against their original sources.

\bibliography{custom}

\appendix

\section{Tool Definitions}
\label{app:tool_definitions}

Our agent interacts with the Memory Bank and raw context through a structured API plus controller-level terminal-action conventions. Below we provide the detailed specifications for each executable tool available in the \textbf{Memory Toolset} ($\mathcal{T}_{mem}$) and the \textbf{Retrieval Toolset} ($\mathcal{T}_{qa}$), followed by the parsing rules for terminal actions. Each item also states the concise action alias used in Section~\ref{sec:formulation}.

\subsection{Memory Toolset \texorpdfstring{($\mathcal{T}_{mem}$)}{(T-mem)}}
This toolset is used during the \textit{Sequential Memory Maintenance} phase to manage the persistent Memory Bank ($\mathcal{B}_t$).

\begin{itemize}[leftmargin=*]
    \item \texttt{Add}: \texttt{memory\_add(title: str, content: str) $\rightarrow$ str}: 
    Creates a new entry in the memory bank. 
    \textbf{Input:} A unique \texttt{title} and the \texttt{content}. 
    \textbf{Output:} Returns "Success" if the title is new; returns an error message if it already exists.
    
    \item \texttt{Update}: \texttt{memory\_update(title: str, content: str[, new\_title: str]) $\rightarrow$ str}: 
    Modifies or renames an existing entry. 
    \textbf{Input:} The target \texttt{title}, replacement \texttt{content}, and optional \texttt{new\_title}. 
    \textbf{Output:} Returns "Success" if the update is applied; returns an error if the title is not found.
    
    \item \texttt{Delete}: \texttt{memory\_delete(title: str) $\rightarrow$ str}: 
    Removes an entry. 
    \textbf{Input:} The target \texttt{title}. 
    \textbf{Output:} Returns "Success" upon deletion; returns an error if the title does not exist.
    
    \item \texttt{Retrieve}: \texttt{memory\_key\_retrieve(title: str) $\rightarrow$ str}: 
    Reads the full content of a specific entry. 
    \textbf{Input:} The target \texttt{title}. 
    \textbf{Output:} Returns the stored \texttt{content} string; returns an error if the title does not exist.
    
    \item \texttt{List}: \texttt{memory\_list([filter: str]) $\rightarrow$ List[str]}: 
    Scans the memory bank. 
    \textbf{Input:} An optional \texttt{filter} string. 
    \textbf{Output:} Returns matching keys (titles), enabling the agent to survey stored topics before performing operations.
\end{itemize}

\subsection{Retrieval Toolset \texorpdfstring{($\mathcal{T}_{qa}$)}{(T-qa)}}
Used during the \textit{Hybrid Retrieval-Augmented QA} phase to retrieve from the raw-context index. Specifically, \texttt{memory\_bm25\_retrieve} and \texttt{memory\_embedding\_retrieve} query raw context chunks, while \texttt{memory\_key\_retrieve} and \texttt{memory\_list} access the structured Memory Bank. Table~\ref{tab:final_qa_prompt} reproduces the original tool prompt verbatim as used in all experiments, including its legacy naming and wording.

\begin{itemize}[leftmargin=*]
    \item \texttt{Embedding}: \texttt{memory\_embedding\_retrieve( query: str[, top\_k: int]) $\rightarrow$ List[str]}: 
    Performs dense retrieval over raw context chunks.
    \textbf{Input:} A search \texttt{query} and optional \texttt{top\_k} (default: 20). 
    \textbf{Mechanism:} Retrieves the top-$k$ context chunks by semantic similarity.
    \textbf{Output:} Ranked results in JSON format.
    
    \item \texttt{BM25}: \texttt{memory\_bm25\_retrieve(query: str[, top\_k: int]) $\rightarrow$ List[str]}: 
    Performs sparse retrieval over raw context chunks.
    \textbf{Input:} A search \texttt{query} and optional \texttt{top\_k} (default: 20). 
    \textbf{Mechanism:} Uses the BM25 algorithm to match keywords.
    \textbf{Output:} Ranked results in JSON format.

\end{itemize}

\subsection{Terminal Actions and Parsing Convention}
\label{app:terminal_actions}
\texttt{UpdateCore} and \texttt{Answer} are terminal actions in the MDP formulation, but they are not implemented as explicit executable tool calls in the tool executor. Instead, the controller interprets the model's final plain-text response according to the current phase:
\begin{itemize}[leftmargin=*]
    \item \textbf{Memory maintenance phase.} If the agent emits a structured memory-tool call, the executor runs the requested tool and returns its observation. If the agent emits no structured tool call and instead outputs plain text, the controller treats that text as an implicit \texttt{UpdateCore} action: the text is committed as the updated core memory $m_t^{core}$, the current chunk-level session terminates, and the environment advances from $c_k$ to $c_{k+1}$.
    \item \textbf{QA phase.} If the agent emits a retrieval or memory-read tool call, the executor runs the tool and returns its observation. If the agent emits no structured tool call and instead outputs plain text, the controller treats that text as the implicit \texttt{Answer(text)} action and terminates the QA session.
\end{itemize}
Thus, \texttt{UpdateCore} and \texttt{Answer(text)} define phase-specific stopping semantics rather than callable API endpoints. Tool-validity rewards are computed over explicit structured tool calls, while these terminal text outputs are evaluated through the corresponding memory-update or answer-outcome signal.

\section{Agent Prompt Templates}
\label{app:agent_prompts}

We provide the complete prompt templates used by UMA-Generalist for memory maintenance and final question answering. Each prompt includes the phase-specific tool-calling system message followed by the corresponding task instruction. The task instruction uses \textit{episodic} and \textit{semantic} memory as instruction-level categories for deciding what information to store; in implementation, both are uniformly stored as key-value entries in the same Memory Bank, without separate memory modules. The full memory-maintenance and final-QA prompts are shown in Tables~\ref{tab:memory_maintenance_prompt} and~\ref{tab:final_qa_prompt}, respectively.

\begin{SaveVerbatim}[breaklines=true,breakanywhere=true,fontsize=\tiny]{MemoryMaintenancePrompt}
<|im_start|>system
# Tools

You may call one or more functions to assist with the user query.

You are provided with function signatures within <tools></tools> XML tags:
<tools>
{"type": "native", "function": {"name": "memory_add", "description": "Store new information in memory with a unique title for later retrieval", "parameters": {"type": "object", "properties": {"title": {"type": "string", "description": "Human-friendly title used to reference this memory entry"}, "content": {"type": "string", "description": "Content to store in memory"}}, "required": ["title", "content"]}}}
{"type": "native", "function": {"name": "memory_update", "description": "Update or rename existing memory entries by title, replacing their content completely", "parameters": {"type": "object", "properties": {"title": {"type": "string", "description": "Existing memory title to update"}, "content": {"type": "string", "description": "New content to persist"}, "new_title": {"type": "string", "description": "Optional new title for the memory entry"}}, "required": ["title", "content"]}}}
{"type": "native", "function": {"name": "memory_delete", "description": "Permanently delete memory entries by their exact title", "parameters": {"type": "object", "properties": {"title": {"type": "string", "description": "Memory title to delete"}}, "required": ["title"]}}}
{"type": "native", "function": {"name": "memory_key_retrieve", "description": "Retrieve the full content of a memory entry using its exact title", "parameters": {"type": "object", "properties": {"title": {"type": "string", "description": "Memory title to retrieve"}}, "required": ["title"]}}}
{"type": "native", "function": {"name": "memory_list", "description": "List stored memory keys. HIGHLY RECOMMENDED: Use the 'filter' parameter to narrow down results.", "parameters": {"type": "object", "properties": {"filter": {"type": "string", "description": "Optional keyword to filter keys. Only keys containing this string will be returned (e.g., 'Transportation', '2024-04')."}}, "required": []}}}
</tools>

For each function call, return a json object with function name and arguments within <tool_call></tool_call> XML tags:
<tool_call>
{"name": <function-name>, "arguments": <args-json-object>}
</tool_call><|im_end|>
<|im_start|>user
You are an expert assistant for memory management.

### Step 1: External Memory Storage (Tool Calls)
- Episodic Memory: record specific events, user actions, friend actions, or assistant actions with timestamps.
- Semantic Memory: record detailed facts, definitions, relationships, labels.
- Always store these details externally via tools like `memory_add`.
- Do NOT store the Core Memory externally -- it is for internal topic reference only.

### Step 2: Core Memory Update
- Core memory should store only high-level themes and main topics of the context.
- Do NOT include detailed facts, numerical data, timestamps, or full event descriptions.
- Summarize in 1-3 short sentences what the context is about:
  - Main subject / theme
  - Overall purpose or intent
  - Any broad category or domain (e.g., technology, sports, finance, etc.)
- Keep it concise, clear, and relevant to understanding "what this context is generally about".
- If the new section changes or adds themes, update the core memory accordingly.
- Avoid redundancy; merge overlapping topics.

<core_memory>
{memory}
</core_memory>

<section>
{chunk}
</section>

Important:
- Final answer must be the updated concise core memory only (plain text, no tool syntax).
- Max length: 1-3 sentences.
- Focus only on WHAT the context is generally about, not details.

Updated concise core memory:
<|im_end|>
\end{SaveVerbatim}

\begin{SaveVerbatim}[breaklines=true,breakanywhere=true,fontsize=\tiny]{FinalQAPrompt}
<|im_start|>system
# Tools

You may call one or more functions to assist with the user query.

You are provided with function signatures within <tools></tools> XML tags:
<tools>
{"type": "native", "function": {"name": "memory_key_retrieve", "description": "Retrieve the full content of a memory entry using its exact title", "parameters": {"type": "object", "properties": {"title": {"type": "string", "description": "Memory title to retrieve"}}, "required": ["title"]}}}
{"type": "native", "function": {"name": "memory_list", "description": "List stored memory keys. HIGHLY RECOMMENDED: Use the 'filter' parameter to narrow down results.", "parameters": {"type": "object", "properties": {"filter": {"type": "string", "description": "Optional keyword to filter keys. Only keys containing this string will be returned (e.g., 'Transportation', '2024-04')."}}, "required": []}}}
{"type": "native", "function": {"name": "memory_bm25_retrieve", "description": "Search memory using keyword matching (BM25). Best for finding documents with specific terms or exact phrases. Returns JSON with ranked results.", "parameters": {"type": "object", "properties": {"query": {"type": "string", "description": "Natural language query used to retrieve relevant memory chunks."}, "top_k": {"type": "integer", "description": "Maximum number of retrieved documents to return. Defaults to 20 if not specified."}}, "required": ["query"]}}}
{"type": "native", "function": {"name": "memory_embedding_retrieve", "description": "Search memory using semantic similarity (embeddings). Best for finding conceptually related content when exact keywords may vary. Returns JSON with ranked results.", "parameters": {"type": "object", "properties": {"query": {"type": "string", "description": "Natural language query used to retrieve relevant memory documents."}, "top_k": {"type": "integer", "description": "Maximum number of documents to return. Defaults to 20 if not specified."}}, "required": ["query"]}}}
</tools>

For each function call, return a json object with function name and arguments within <tool_call></tool_call> XML tags:
<tool_call>
{"name": <function-name>, "arguments": <args-json-object>}
</tool_call><|im_end|>
<|im_start|>user
You are presented with a problem and a previous memory.

You have access to tools that can help you solve the problem. Use them if necessary. In particular, if you can't find the answer in the memory, you should use the memory_list, memory_key_retrieve, memory_embedding_retrieve and memory_bm25_retrieve tools to find relevant information. If you can not find the answer using one of these tools, you can also use other tools to help you. Try as hard as you can to find the answer using the tools available to you until you find the answer or you are sure that the answer is not in the memory. Do not decide that the answer is not in the memory unless you have used all relevant tools to search for the answer. Before deciding that the information is unavailable, make sure you have used all relevant memory and retrieval tools.


Please answer the problem based on the previous memory and the tools. Put your final answer in \boxed{{}}.

<problem>
{prompttext}
</problem>

<memory>
{memory}
</memory>

Your answer:
<|im_end|>
\end{SaveVerbatim}

\begin{table*}[!b]
    \centering
    \setlength{\tabcolsep}{0pt}
    \renewcommand{\arraystretch}{1.05}
    \begin{tabular}{@{}p{0.98\textwidth}@{}}
    \toprule
    \textbf{Memory Maintenance Prompt} \\
    \midrule
    \begin{minipage}[t]{0.98\textwidth}
    \UseVerbatim[breaklines=true,breakanywhere=true,fontsize=\tiny]{MemoryMaintenancePrompt}
    \end{minipage} \\
    \bottomrule
    \end{tabular}
    \caption{Complete prompt template for UMA memory maintenance. The table spans both columns and includes both the tool-calling system message and task instruction.}
    \label{tab:memory_maintenance_prompt}
\end{table*}

\begin{table*}[!b]
    \centering
    \setlength{\tabcolsep}{0pt}
    \renewcommand{\arraystretch}{1.05}
    \begin{tabular}{@{}p{0.98\textwidth}@{}}
    \toprule
    \textbf{Final QA Prompt} \\
    \midrule
    \begin{minipage}[t]{0.98\textwidth}
    \UseVerbatim[breaklines=true,breakanywhere=true,fontsize=\tiny]{FinalQAPrompt}
    \end{minipage} \\
    \bottomrule
    \end{tabular}
    \caption{Complete prompt template for UMA final question answering. The table spans both columns and includes both the tool-calling system message and task instruction.}
    \label{tab:final_qa_prompt}
\end{table*}

\section{Details of Ledger-QA Construction}
\label{app:ledger_construction}

The construction of the Ledger-QA benchmark employs a three-stage pipeline designed to simulate realistic, long-horizon bookkeeping scenarios. In our implementation, we adopt a \textbf{joint synthesis strategy}: we provide the LLM with high-level user intents (e.g., "wants to buy coffee"), and the LLM simultaneously generates both the natural dialogue and the corresponding structured transaction records. This design helps ensure that details such as amounts and descriptions are contextually appropriate and naturally woven into the conversation.

\subsection{Consumption Scene Taxonomy}
To simulate a diverse personal finance environment, we define a hierarchical taxonomy of consumption scenes. The dataset covers 8 major categories, each with multiple sub-scenes:
\begin{itemize}[nosep]
    \item \textbf{Dining:} Fast Food, Restaurant, Coffee, Bubble Tea, BBQ, Hot Pot, Snacks, Takeout.
    \item \textbf{Transportation:} Subway, Bus, Taxi, Gas, Parking, Train, Flight.
    \item \textbf{Shopping:} Clothing, Electronics, Daily Necessities, Cosmetics, Books, Groceries, Furniture.
    \item \textbf{Entertainment:} Movie, KTV, Gaming, Gym, Travel, Concert, Escape Room.
    \item \textbf{Utilities:} Water \& Electricity, Property Fee, Phone Bill, Internet, Gas Bill, Rent.
    \item \textbf{Medical:} Medicine, Doctor Visit, Health Checkup, Dental, Glasses.
    \item \textbf{Education:} Training Course, Books \& Materials, Online Course, Exam Registration, Tuition.
    \item \textbf{Other:} Transfer, Red Envelope, Donation, Pet, Beauty \& Salon.
\end{itemize}

\subsection{Data Generation Pipeline}

\paragraph{Stage 1: Timeline \& Intent Sampling.}
For each sample, we randomly sample $H$ dates (sessions) from \texttt{2023-01-01} to \texttt{2024-12-31}. For each session, we stochastically select a subset of consumption scenes (e.g., \textit{Dining-Coffee}, \textit{Transportation-Taxi}) to form the user's "intent" for that day.

\paragraph{Stage 2: Joint Dialogue \& Transaction Synthesis.}
We feed the selected date and scene intents into an advanced LLM (e.g., Gemini 3 Pro~\citep{pichai2025gemini3}) to obtain natural language realizations while keeping generation controls explicit. The model is instructed to act as a user-assistant pair and generate a JSON object containing two aligned fields:
\begin{enumerate}
    \item \textbf{Dialogue:} The natural conversation turns.
    \item \textbf{Transactions:} The structured ground truth (Amount, Description, etc.) corresponding to the dialogue.
\end{enumerate}
Crucially, we do not pre-determine the amounts. The LLM decides reasonable amounts based on the context (e.g., \$5 for coffee vs. \$500 for a flight), ensuring semantic consistency. To validate the dialogue--transaction alignment, we manually audited 100 transactions randomly sampled from the Stage-2 outputs. All 100 sampled transactions were accurately supported by the corresponding dialogue, and their scenes and amounts were judged contextually reasonable, indicating that the LLM-generated dialogues and structured transactions are reliable for this synthesis step.

\begin{table}[h]
    \centering
    \small
    \begin{tabular}{@{}p{0.95\linewidth}@{}}
    \toprule
    \textbf{Prompt for Joint Synthesis (Simplified)} \\
    \midrule
    Please generate a natural conversation between a user and an AI assistant about expense tracking. \\
    \textbf{Context:} \\
    - Date: \{date\} \\
    - The user wants to record: \{selected\_scenes\} \\
    - \textit{[Instruction: First Session vs. Continuation]} \\
    \textbf{Requirements:} \\
    - The dialogue should be natural, containing \{min\_turns\}-\{max\_turns\} turns. \\
    - Each consumption scene needs a specific amount (\textbf{you decide a reasonable amount}). \\
    - Return JSON with two fields: \\
      1. "dialogue": List of turns. \\
      2. "transactions": List of records (scene, amount, description, date). \\
    \bottomrule
    \end{tabular}
    \caption{Prompt template used for the joint generation of dialogue and structured data.}
    \label{tab:prompt_dialogue}
\end{table}

\paragraph{Example Multi-Session Sample from Stage 2 (Excerpt).}
Table~\ref{tab:stage2_example} shows a shortened multi-session sample generated by the Stage-2 joint synthesis pipeline. We include two sessions from different dates, aligned transaction records with explicit date fields, and key metadata, to illustrate the temporal structure used by Ledger-QA.

\begin{table}[h]
    \centering
    \scriptsize
    \begin{tabular}{@{}p{0.95\linewidth}@{}}
    \toprule
    \textbf{Session Dates:} 2024-02-23 and 2024-04-29 \\
    \midrule
    \textbf{Dialogue Excerpts (cross-date)} \\
    \texttt{[2024-02-23][User]} Hi there! I'd like to start using this chat to track my daily expenses. \\
    \texttt{[2024-02-23][Assistant]} Great initiative. I can help log each expense with category and amount. \\
    \texttt{[2024-02-23][User]} I paid my monthly Internet bill. It cost 75. \\
    \texttt{[2024-02-23][Assistant]} Got it. Should I categorize this under Utilities? \\
    \texttt{[2024-02-23][User]} Yes, please put it under Utilities. \\
    \texttt{[2024-02-23][Assistant]} Recorded: Utilities--Internet, 75. Do you have other expenses? \\
    \texttt{\ldots} \\
    \texttt{[2024-04-29][User]} I have a few more receipts. First, I took a cab to the city center. \\
    \texttt{[2024-04-29][Assistant]} Transportation--Taxi noted. How much was the fare? \\
    \texttt{[2024-04-29][User]} 25.5. Also, a clinic co-pay (50), a gas bill (85.20), and skincare products (120). \\
    \texttt{[2024-04-29][Assistant]} Done. I've saved all four transactions for April 29th. \\
    \texttt{\ldots} \\
    \midrule
    \textbf{Aligned Structured Transactions (date, scene, subscene, amount)} \\
    1) (2024-02-23, Utilities, Internet, 75.00) \\
    2) (2024-02-23, Transportation, Subway, 2.90) \\
    3) (2024-02-23, Entertainment, Escape Room, 40.00) \\
    4) (2024-02-23, Education, Tuition, 1500.00) \\
    5) (2024-02-23, Transportation, Bus, 1.50) \\
    6) (2024-04-29, Transportation, Taxi, 25.50) \\
    7) (2024-04-29, Medical, Doctor Visit, 50.00) \\
    8) (2024-04-29, Utilities, Gas Bill, 85.20) \\
    9) (2024-04-29, Shopping, Cosmetics, 120.00) \\
    \midrule
    \textbf{Sample Metadata} \\
    num\_sessions=2; num\_turns=48; num\_transactions=9; \\ date\_range=[2024-02-23, 2024-04-29]; total\_amount=1900.10. \\
    \bottomrule
    \end{tabular}
    \caption{Shortened multi-session Stage-2 generation example. The excerpt preserves cross-date dialogue evidence, date-aware transaction records, and core metadata.}
    \label{tab:stage2_example}
\end{table}

\paragraph{Stage 3: Ground-Truth Query Generation.}
From the transaction ledger generated in Stage 2, we programmatically generate questions and exact answers across 8 distinct templates.

The question types include:
\begin{enumerate}
    \item \textbf{Time-Range Scene Amount:} "How much was spent on Dining from Jan to Mar?"
    \item \textbf{Time-Range Multi-Scene:} "Total spending on Dining and Travel in Q1?"
    \item \textbf{Global Total:} "What was the total spending across all records?"
    \item \textbf{Max Scene:} "Which category had the highest spending?"
    \item \textbf{Max Frequency Date:} "On which date were there the most transactions?"
    \item \textbf{Max Single Amount:} "What was the largest single transaction?"
    \item \textbf{Point Query:} "How much was spent on Coffee on 2024-05-01?"
    \item \textbf{Single Date Scene Amount:} "How much was spent on Dining on 2024-05-01?"
\end{enumerate}

\subsection{Dataset Statistics}
To evaluate the agent's robustness across varying temporal horizons, we constructed a comprehensive suite of datasets ranging from short-term interactions (2 sessions) to ultra-long histories (500 sessions). 

\paragraph{Training Set.} We utilize the $H=10$ configuration for training, consisting of 200 samples with over 6,000 questions. This provides a balanced difficulty level for the agent to learn state-tracking policies.

\paragraph{Test Set.} The test set is stratified by session horizon ($H \in \{2, 5, 10, 20, 30, 40, 50, 100, 500\}$). Detailed statistics for each configuration are presented in Table~\ref{tab:all_data_stats}.

\subsection{Generation Algorithm}
The core logic of our joint synthesis pipeline is formalized in Algorithm~\ref{alg:ledger_gen}.

\begin{algorithm*}[h]
\caption{Ledger-QA Data Generation Pipeline}
\label{alg:ledger_gen}
\begin{algorithmic}[1]
\Require Date range $[D_{start}, D_{end}]=[\texttt{2023-01-01}, \texttt{2024-12-31}]$, Session horizon $H$, Diversify flag $diversify$
\Ensure Context chunks $\mathcal{C}$, Question set $\mathcal{Q}$, Ground-truth Ledger $\mathcal{L}$

\State $\mathcal{D} \leftarrow \text{SampleSortedDates}(D_{start}, D_{end}, H)$
\State $\mathcal{L} \leftarrow \emptyset$ \Comment{Initialize global transaction ledger}
\State $\mathcal{C} \leftarrow \emptyset$ \Comment{Initialize dialogue history}

\For{$i \leftarrow 1$ to $H$}
    \State $d_i \leftarrow \mathcal{D}[i]$
    \State $\mathcal{S}_{intent} \leftarrow \text{SampleScenes}(\text{Taxonomy})$ \Comment{Randomly select user intents}
    
    \State $P \leftarrow \text{FormatPrompt}(d_i, \mathcal{S}_{intent}, \text{is\_first}=(i==1))$ \Comment{Construct prompt for joint synthesis}
    
    \State $\text{Response} \leftarrow \text{LLM}_{\text{gen}}(P)$ \Comment{Generate dialogue and transaction simultaneously}
    \State $\text{Dialogue}_i, \text{Trans}_i \leftarrow \text{ParseJSON}(\text{Response})$
    
    \State $\mathcal{C}.\text{append}(\text{FormatChunk}(d_i, \text{Dialogue}_i))$
    \State $\mathcal{L}.\text{extend}(\text{Trans}_i)$
\EndFor

\State $\mathcal{Q}_{raw} \leftarrow \text{ApplyTemplates}(\mathcal{L}, \mathrm{Types}=\{1,\dots,8\})$ \Comment{Generate questions based on the full ledger}

\If{diversify} \Comment{Optional: Diversify phrasing}
    \State $\mathcal{Q} \leftarrow \text{LLM}_{\text{paraphrase}}(\mathcal{Q}_{raw})$
\Else
    \State $\mathcal{Q} \leftarrow \mathcal{Q}_{raw}$
\EndIf

\State \Return $\mathcal{C}, \mathcal{Q}, \mathcal{L}$
\end{algorithmic}
\end{algorithm*}

\section{Evaluation Prompts}
\label{app:eval_prompts}

To ensure consistent and fair evaluation, we employ an LLM-as-a-Judge protocol with two independent evaluators: \texttt{DeepSeek-V3.2} (main results) and \texttt{Qwen3-30B-A3B-Instruct-2507} (supplementary comparison). The specific prompts used for different benchmarks are detailed below.

\paragraph{General Prompt.}
For all benchmarks except LongMemEval, we utilize a unified evaluation template designed to assess semantic correctness:

\begin{quote}
\small
"I will give you a question, a correct answer, and a response from a model. Please answer yes if the response contains the correct answer. Otherwise, answer no. If the response is equivalent to the correct answer or contains all the intermediate steps to get the correct answer, you should also answer yes. If the response only contains a subset of the information required by the answer, answer no.

\textbf{Question:} \{question\}

\textbf{Correct Answer:} \{answer\}

\textbf{Model Response:} \{prediction\}

Is the model response correct? Answer yes or no only."
\end{quote}

\paragraph{LongMemEval Prompts.}
Following the official implementation~\citep{wulongmemeval}, we apply task-specific prompts to handle the nuances of different reasoning types (e.g., temporal reasoning, knowledge updates).

\begin{itemize}[leftmargin=*]
    \item \textbf{Default / Fact Retrieval:} Same as the General Prompt above.
    
    \item \textbf{Temporal Reasoning:} Adds tolerance for off-by-one errors:
    \begin{quote}
    \small
    "...In addition, do not penalize off-by-one errors for the number of days. If the question asks for the number of days/weeks/months, etc., and the model makes off-by-one errors (e.g., predicting 19 days when the answer is 18), the model's response is still correct..."
    \end{quote}
    
    \item \textbf{Knowledge Update:} Allows inclusion of previous information if the update is correct:
    \begin{quote}
    \small
    "...If the response contains some previous information along with an updated answer, the response should be considered as correct as long as the updated answer is the required answer..."
    \end{quote}
    
    \item \textbf{Single-Session Preference:} Evaluates personalization based on a rubric:
    \begin{quote}
    \small
    "...Please answer yes if the response satisfies the desired response... The response is correct as long as it recalls and utilizes the user's personal information correctly.
    
    \textbf{Rubric:} \{rubric\}..."
    \end{quote}
    
    \item \textbf{Abstention:} Checks for correct identification of unanswerable questions:
    \begin{quote}
    \small
    "...Please answer yes if the model correctly identifies the question as unanswerable. The model could say that the information is incomplete...
    
    \textbf{Explanation:} \{explanation\}
    
    Does the model correctly identify the question as unanswerable? Answer yes or no only."
    \end{quote}
\end{itemize}

\section{Evaluation Dataset Details}
\label{app:data_details}

\begin{table*}[t]
    \centering
    \small
    \renewcommand{\arraystretch}{1.1}
    \begin{tabular}{l l r r r r}
    \toprule
    \textbf{Category} & \textbf{Dataset} & \textbf{Samples} & \textbf{Avg. Qs} & \textbf{Avg. Len (Tok)} & \textbf{Avg. Chunks} \\
    \midrule
    \multirow{9}{*}{State Tracking} 
    & Ledger-QA ($H=2$) & 5 & 20.6 & 1,693 & 2.0 \\
    & Ledger-QA ($H=5$) & 4 & 30.8 & 4,204 & 5.0 \\
    & Ledger-QA ($H=10$) & 3 & 34.3 & 8,192 & 10.0 \\
    & Ledger-QA ($H=20$) & 3 & 34.0 & 16,487 & 20.0 \\
    & Ledger-QA ($H=30$) & 3 & 34.3 & 24,772 & 30.0 \\
    & Ledger-QA ($H=40$) & 3 & 35.0 & 32,792 & 40.0 \\
    & Ledger-QA ($H=50$) & 3 & 35.0 & 40,610 & 50.0 \\
    & Ledger-QA ($H=100$) & 3 & 34.7 & 81,587 & 100.0 \\
    & Ledger-QA ($H=500$) & 3 & 34.7 & 405,990 & 500.0 \\
    \midrule
    \multirow{6}{*}{Test Time Learning} 
    & Banking77 & 1 & 100.0 & 127,568 & 111.0 \\
    & Clinic & 1 & 100.0 & 130,702 & 38.0 \\
    & NLU & 1 & 100.0 & 134,060 & 115.0 \\
    & PubMed-RCT & 10 & 100.0 & 16,724 & 10.0 \\
    & TREC-Coarse & 1 & 100.0 & 123,606 & 111.0 \\
    & TREC-Fine & 1 & 100.0 & 125,531 & 108.0 \\
    \midrule
    \multirow{7}{*}{Accurate Retrieval} 
    & ConvoMem & 7 & 64.0 & \textbf{522,602} & 300.0 \\
    & HotpotQA & 128 & 1.0 & 27,945 & 200.0 \\
    & LoCoMo & 10 & 198.6 & 22,161 & 27.2 \\
    & LongMemEval & 500 & 1.0 & 107,863 & 47.8 \\
    & MSC (Batch=5) & 100 & 5.0 & 9,853 & 25.0 \\
    & PerLTQA & 4 & 100.0 & 13,038 & 23.0 \\
    & SQuAD & 30 & 96.8 & 10,561 & 10.0 \\
    \bottomrule
    \end{tabular}
    \caption{Detailed statistics of all evaluation datasets. \textbf{Avg. Len (Tok)} denotes the average total context length in tokens. \textbf{Avg. Chunks} denotes the average number of context chunks per sample processed by the agent.}
    \label{tab:all_data_stats}
\end{table*}

We employ a diverse set of benchmarks to evaluate the agent's capabilities. Detailed statistics for all evaluation datasets are summarized in Table~\ref{tab:all_data_stats}. Below we detail the preprocessing and construction logic for each dataset, corresponding to the implementation in our codebase.

\paragraph{Artifact Licenses and Intended Use.}
We use third-party artifacts under the licenses, access terms, and documentation provided by their original releases. LoCoMo, PerLTQA, and ConvoMem are used only for non-commercial research under CC BY-NC 4.0. The release pages for the Mem-$\alpha$ corpus and the Mem-$\alpha$/Mem1 checkpoints do not currently declare licenses; we therefore cite them through their original sources and do not redistribute these artifacts. Ledger-QA and the arithmetic-control set are synthetic artifacts created for this study.

\subsection{Dynamic State Tracking}
\paragraph{Ledger-QA (Synthetic).}
As detailed in Appendix~\ref{app:ledger_construction}, we generate synthetic bookkeeping data. For evaluation, we use a stratified test set across session horizons $H \in \{2, 5, 10, 20, 30, 40, 50, 100, 500\}$, as summarized in Table~\ref{tab:all_data_stats}.

\paragraph{Arithmetic-Control Set.}
We construct 100,000 arithmetic-only examples to isolate numerical computation: each asks for the sum of five randomly sampled numbers below 1000 rounded to two decimals. Exact-match accuracy is computed against the generated sum. The set contains no long-context stream, memory updates, or retrieval requirement.

\subsection{Test-Time Learning (TTL)}
For TTL tasks, we source data from \textbf{MemoryAgentBench}~\citep{hu2025evaluating}. To ensure the agent understands the classification nature of these tasks, we prepend a specific instruction hint to each query if not already present:
\begin{quote}
\textit{"Sentence: \{original\_query\} \\ What are the labels for the above sentence?"}
\end{quote}

\subsection{Accurate Retrieval (AR)}

\paragraph{HotpotQA.} Following MemAgent~\citep{yu2025memagent}, we ensure strict data isolation by using non-overlapping training and validation splits. We inject ground-truth documents into 200 shuffled distractors, requiring the agent to retrieve evidence from the chunked stream to solve multi-hop queries without any risk of data leakage.

\paragraph{LoCoMo.}
We adapt prompts for specific question categories to ensure valid evaluation:
\begin{itemize}[nosep]
    \item \textbf{Temporal (Cat 2):} We append \textit{"Use DATE of CONVERSATION to answer with an approximate date"} to guide relative time reasoning.
    \item \textbf{Adversarial (Cat 5):} We convert these into a multiple-choice format \textit{"(A) \{distractor\} (B) Not mentioned"} (with randomized order). The agent must select \textit{"Not mentioned"} to demonstrate correct rejection of unanswerable queries.
\end{itemize}

\paragraph{LongMemEval.}
We utilize the \texttt{longmemeval\_s} subset. Questions are timestamped (e.g., \texttt{[2024-01-01] Question\dots}) to provide temporal context.

\paragraph{MSC (Multi-Session Chat).}
We use a modified version of MSC introduced by MemGPT~\citep{packer2023memgpt} provided at \url{https://huggingface.co/datasets/MemGPT/MSC-Self-Instruct}. To increase context length, we batch 5 base samples into a single evaluation instance (\texttt{batch\_size=5}). Each instance thus contains 25 sessions (5 samples $\times$ 5 sessions), and the agent is queried about specific details from any of the batched conversations.

\paragraph{ConvoMem.}

We curate evaluation samples from the \texttt{pre\_mixed\_testcases} split provided at \url{https://huggingface.co/datasets/Salesforce/ConvoMem/tree/main}, retaining only conversations whose total length exceeds 2 million characters to ensure they represent genuine long-context tasks. Each conversation turn is formatted as "\{Speaker\}: \{Text\}". We use the \texttt{user\_evidence} category for evaluation.

\section{Baseline Implementation Details}
\label{app:base_details}

We implement the baseline methods \textbf{Mem1} and \textbf{Mem-$\alpha$} using the codebase provided at \url{https://github.com/wangyu-ustc/Mem-alpha}. Since \textbf{MemAgent} follows a simple recurrent summarization procedure, we implement it ourselves while using the official prompt templates. We implement \textbf{A-MEM} from the authors' reproduction repository at \url{https://github.com/WujiangXu/A-mem}, retaining its released memory-construction and evolution prompts and default retrieval/update configuration; final answers use our unified QA prompt and evaluator. We implement \textbf{Mem-T} using the official open-source model \texttt{EdwinYue/Mem-T-4B} and the core implementation from \url{https://github.com/yanweiyue/Mem-T/tree/main}, with only interface-level adaptations to integrate all baselines into our chunked-stream evaluation pipeline and unified evaluator. We keep released memory-formation checkpoints, prompts, and default inference hyperparameters whenever available, and do not tune baseline hyperparameters on the test sets. In evaluation, \emph{context budget} denotes the cumulative token budget of one multi-turn agent session. It covers the initial prompt (instructions, memory or raw context, and query) together with all subsequent assistant actions, tool observations, and interaction history. We generally allocate half of the budget to the initial prompt and reserve the other half for the remaining multi-turn interaction. Unless otherwise specified, the total session budget is 16k tokens; the 8k/32k variants change this cap, while the latency-only Concat-256k configuration uses 256k. We left-truncate the memory or raw-context portion when the initial prompt exceeds its allocation, while preserving instructions and the query.

For a fair comparison, all embedding-based retrieval components use the same dense encoder, \texttt{sentence-transformers/all-MiniLM-L6-v2}. This applies to RAG and to memory-agent baselines with embedding retrieval components, including A-MEM, Mem-$\alpha$, and Mem-T.

\paragraph{Backbone Models.}
To isolate the contribution of the memory mechanism, we standardize the backbone model where possible:
\begin{itemize}[leftmargin=*]
    \item \textbf{Standard Backbone:} Concat and RAG use \texttt{Qwen/Qwen3-4B-Instruct-2507} for answer generation. MemAgent and MemAgent-woq use the same model for recurrent memory updates and final QA. A-MEM uses it for memory construction and evolution as well as final QA.
    
    \item \textbf{Specialized Memory Models:} For baselines that require specific pre-trained weights for memory operations, we follow their original configurations:
    \begin{itemize}
        \item \textbf{Mem1:} Uses \path{Mem-Lab/Qwen2.5-7B-RL-RAG-Q2-EM-Release} for the memory formation phase.
        \item \textbf{Mem-$\alpha$:} Uses \texttt{YuWangX/Memalpha-4B} for the memory formation phase.
        \item \textbf{Mem-T:} Uses the official \texttt{EdwinYue/Mem-T-4B} checkpoint. Since Mem-T designs speaker-aware memories for dialogue, for benchmarks that are not dialogue-formatted, we treat each complete chunk as a single turn and set the speaker to \texttt{context}.
    \end{itemize}
    Crucially, for the final Question Answering phase, both Mem1 and Mem-$\alpha$ use the standard \texttt{Qwen/Qwen3-4B-Instruct-2507} backbone. The original Mem-$\alpha$ repository uses a larger Qwen3-32B QA model; for fair comparison under our evaluation setting, we replace only this QA model and keep the released Mem-$\alpha$ memory-formation model unchanged.
\end{itemize}

\section{Comparison of Trainable Memory Agents}
\label{app:memory_agent_comparison}

Table~\ref{tab:memory_agent_comparison} compares five training dimensions among closely related memory agents. UMA combines query-hidden structured memory, shared-memory supervision from multiple questions, joint memory--QA policy training, and stratified advantage groups.

\begin{table}[H]
    \centering
    \scriptsize
    \setlength{\tabcolsep}{2.5pt}
    \renewcommand{\arraystretch}{1.12}
    \resizebox{\columnwidth}{!}{
    \begin{tabular}{@{}lccccc@{}}
        \toprule
        \textbf{Method} &
        \shortstack{\textbf{Question}\\\textbf{hidden}} &
        \shortstack{\textbf{Structured}\\\textbf{bank}} &
        \shortstack{\textbf{Shared-memory}\\\textbf{multi-Q}} &
        \shortstack{\textbf{Joint memory--}\\\textbf{QA policy}} &
        \shortstack{\textbf{Stratified}\\\textbf{groups}} \\
        \midrule
        A-MEM        & \checkmark & \checkmark & --         & --         & -- \\
        MemAgent     & --         & --         & --         & \checkmark & -- \\
        ReMemR1      & --         & --         & --         & \checkmark & -- \\
        Mem-$\alpha$ & \checkmark & \checkmark & \checkmark & --         & -- \\
        Memory-R1    & \checkmark & \checkmark & --         & --         & -- \\
        Mem-T        & \checkmark & \checkmark & --         & --         & -- \\
        AgeMem       & \checkmark & \checkmark & --         & \checkmark & -- \\
        \textbf{UMA} & \checkmark & \checkmark & \checkmark & \checkmark & \checkmark \\
        \bottomrule
    \end{tabular}
    }
    \caption{Comparison of closely related memory-agent training formulations. ``Question hidden'' means that memory is constructed before the target question is exposed. ``Shared-memory multi-Q'' means that multiple distinct QA trajectories supervise the same sampled memory state. ``Joint memory--QA policy'' means that one trainable policy receives updates from both stages. ``Stratified groups'' means that memory and per-question QA rollouts use separate advantage-normalization groups.}
    \label{tab:memory_agent_comparison}
\end{table}

\section{Training Data Details}
\label{app:train_data_details}

Table~\ref{tab:train_data_stats} summarizes the detailed statistics of the datasets utilized in our training phase. 

\paragraph{Data Sources.}
Our training data is curated from three main sources:
(1) \textbf{HotpotQA (Modified):} We construct a retrieval-intensive version of HotpotQA following the protocol in MemAgent~\citep{yu2025memagent}, augmenting each sample with 200 documents.
(2) \textbf{Ledger-QA (Synthetic):} We generate a specialized dataset for dynamic state tracking as described in Section~\ref{sec:ledger_qa}.
(3) \textbf{Mem-$\alpha$ Corpus:} For all other datasets (SQuAD, PerLTQA, NLU, TREC, PubMed, BookSum), we directly incorporate the high-quality training corpus provided by Mem-$\alpha$~\citep{wang2025mem} to ensure broad coverage of reasoning tasks (provided at \url{https://huggingface.co/datasets/YuWangX/Memalpha-full}).

\paragraph{Data Isolation.}
We strictly enforce data isolation to prevent leakage. For \textbf{HotpotQA}, we follow the MemAgent protocol, constructing our training and evaluation sets from the original HotpotQA training and validation splits, respectively. For \textbf{Ledger-QA}, the training and test samples are synthesized independently with distinct seed parameters. For the \textbf{Mem-$\alpha$ Corpus}, we start from the official training split provided by the authors and remove the LongMemEval portion to avoid evaluation-data leakage. We also remove the HotpotQA portion, both to avoid leakage and because our training data already includes the MemAgent-style modified HotpotQA data described above.

To maintain training efficiency, for Mem-$\alpha$ corpus containing a large number of queries per instance (e.g., SQuAD, NLU), we randomly sample up to 10 queries per training sample.

\begin{table*}[t]
    \centering
    \small
    \renewcommand{\arraystretch}{1.1}
    \begin{tabular}{l r r r r}
    \toprule
    \textbf{Dataset} & \textbf{Samples} & \textbf{Avg. Qs} & \textbf{Avg. Len (Tok)} & \textbf{Avg. Chunks} \\
    \midrule
    HotpotQA & 8,192 & 1.0 & 25,667 & 61 \\
    SQuAD & 264 & 10.0 & 10,780 & 10.0 \\
    PerLTQA & 27 & 10.0 & 12,046 & 23.3 \\
    NLU & 180 & 10.0 & 6,100 & 10.0 \\
    TREC-Coarse & 180 & 10.0 & 3,900 & 10.0 \\
    PubMed-RCT & 90 & 10.0 & 16,760 & 10.0 \\
    BookSum & 1,387 & 1.0 & 15,328 & 8.0 \\
    Ledger-QA & 200 & 34.0 & 8,254 & 10.0 \\
    \bottomrule
    \end{tabular}
    \caption{Detailed statistics of the training datasets. \textbf{Avg. Len (Tok)} denotes the average total context length in tokens. \textbf{Avg. Chunks} denotes the average number of context chunks per sample processed by the agent.}
    \label{tab:train_data_stats}
\end{table*}

\section{Training Configuration}
\label{app:train_config}

We implement our training framework based on a customized version of \texttt{veRL}~\citep{sheng2025hybridflow}, optimized for multi-conversation and multi-turn agent rollouts. The policy model is initialized from \texttt{Qwen/Qwen3-4B-Instruct-2507} and optimized using AdamW with a learning rate of $1\times 10^{-6}$. We employ the Task-Stratified Group Relative Policy Optimization (GRPO) algorithm with a group size of $G=16$ samples per context, reward weight $\lambda=0.5$, and KL divergence penalty coefficient $\beta=0.001$.

To balance generalization across diverse tasks with the specific demands of intensive state tracking, we adopt a dual-track training strategy:

\begin{itemize}[leftmargin=*]
    \item \textbf{UMA-Generalist (for TTL \& AR):} This checkpoint is trained for 2 epochs on the modified HotpotQA set and the filtered Mem-$\alpha$ training split listed in Table~\ref{tab:train_data_stats}. The HotpotQA and LongMemEval portions of the released Mem-$\alpha$ corpus are excluded as described above.
    \item \textbf{UMA-Specialist (for Ledger-QA):} This checkpoint is trained for 6 epochs on the independently generated Ledger-QA training set.
\end{itemize}

All training runs are conducted on a cluster of 32 NVIDIA H200 GPUs (4 nodes $\times$ 8 GPUs). To manage memory efficiency during long-horizon rollouts, we use vLLM with a tensor parallelism size of 1 and set GPU memory utilization to 0.5. The initial prompt and the remaining multi-turn interaction are each allocated up to 8,192 tokens, giving a maximum total session length of 16,384 tokens. The latter allocation is shared cumulatively across all generated actions, tool observations, and later-turn inputs and outputs over the complete session. The 8k/32k variants change this total session cap at inference. Training takes approximately 2 days for UMA-Generalist and 0.5 days for UMA-Specialist.

\section{Inference Efficiency and Resource Scalability}
\label{app:efficiency_scalability}

We benchmark inference latency on a single ConvoMem context containing 528k tokens using 8 NVIDIA A800 GPUs. We randomly sample 10 queries from the selected context for this latency evaluation. Results are shown in Table~\ref{tab:inference_latency}. UMA-Generalist reaches a practical sweet spot: it is substantially faster than recurrent baselines while preserving strong long-horizon reasoning capability.

\begin{table*}[t]
    \centering
    \small
    \setlength{\tabcolsep}{5pt}
    \renewcommand{\arraystretch}{1.15}
    \begin{tabular}{lcc}
    \toprule
    \textbf{Method} & \textbf{Latency (s)} & \textbf{Relative Speed} \\
    \midrule
    \textbf{UMA-Generalist} & \textbf{107.32} & \textbf{1.0x} \\
    Mem1 & 368.46 & 3.4x Slower \\
    A-MEM & 3251.75 & 30.3x Slower \\
    Mem-$\alpha$ & 1503.87 & 14.0x Slower \\
    MemAgent-woq & 2718.88 & 25.3x Slower \\
    MemAgent & 4753.26 & 44.3x Slower \\
    Mem-T & 8611.18 & 80.2x Slower \\
    Concat & 2.58 & 41.6x Faster \\
    Concat-256k & 61.16 & 1.8x Faster \\
    RAG & 58.01 & 1.9x Faster \\
    \bottomrule
    \end{tabular}
    \caption{Inference latency comparison on a single 528k-token ConvoMem context with 10 randomly sampled queries, using 8 NVIDIA A800 GPUs. The UMA result uses UMA-Generalist.}
    \label{tab:inference_latency}
\end{table*}

The results indicate that UMA-Generalist achieves a favorable efficiency--capability balance. Compared with MemAgent, UMA-Generalist processes the full context only once during memory construction and then answers subsequent queries using the maintained memory together with lightweight retrieval. MemAgent traverses the context separately for each query, leading to substantially higher latency when multiple queries are asked over the same long context. Mem-T is the slowest method in this comparison, which may be attributed to its more complex memory architecture, workflow, and retrieval strategy.

\paragraph{Ledger-QA Scaling.}
We separately profile UMA-Specialist over 30 independent context runs on 8 NVIDIA A800 GPUs. Memory construction processes all chunks once and grows approximately linearly with input scale ($R^2=0.995$). Complete QA-stage wall time remains broadly stable from $H=20$ to $H=500$, where each context contains about 35 questions; representative horizon means are shown in Table~\ref{tab:ledger_efficiency}. The reported group-level wall time covers concurrently executed questions (concurrency limit 10), answer generation, memory/raw-context retrieval, and retrieval-index construction. Offline judge scoring is excluded.

\begin{table}[t]
    \centering
    \small
    \setlength{\tabcolsep}{4pt}
    \renewcommand{\arraystretch}{1.05}
    \begin{tabular}{rccc}
    \toprule
    $H$ & \textbf{Avg. Qs} & \textbf{Construct (s)} & \textbf{QA Stage (s)} \\
    \midrule
    20  & 34.0 & 21.15  & 18.25 \\
    40  & 35.0 & 32.09  & 22.85 \\
    50  & 35.0 & 43.16  & 17.65 \\
    100 & 34.7 & 80.78  & 19.28 \\
    500 & 34.7 & 564.67 & 21.55 \\
    \bottomrule
    \end{tabular}
    \caption{Mean Ledger-QA wall time per context for UMA-Specialist. Construction is performed once; the QA stage covers the complete concurrent question group.}
    \label{tab:ledger_efficiency}
\end{table}

The final serialized Memory Bank grows from 18.4 entries (1.15 KB) at $H=2$ to 4,452.7 entries (283.96 KB) at $H=500$. At $H=500$, its mean serialized size is approximately 24.3\% of the raw-context size. Moreover, \texttt{Update} accounts for 32.2\% of all \texttt{Add}/\texttt{Update} writes, showing that the agent revises existing state rather than only appending new entries.

\paragraph{Resource Scalability.}
Beyond large-cluster training, we verify that the training pipeline is compatible with smaller setups and can run on a single node with 8 NVIDIA A800 GPUs (80GB HBM each). Moreover, parameter-efficient fine-tuning methods (e.g., LoRA) offer a practical path to further reduce memory requirements, improving accessibility for broader research groups.

\section{Detailed Evaluation Metrics}
\label{app:detailed_metrics}

In addition to the main LLM-as-a-Judge results, we report deterministic lexical metrics using the same pipeline for all methods. If present, the final answer is first extracted from \texttt{\textbackslash boxed\{...\}} or \texttt{\textbackslash fbox\{...\}}. We then lowercase the answer, directly delete all ASCII punctuation, and collapse consecutive whitespace. Articles are retained, and we apply no numeric, Unicode, morphological, or semantic normalization. Because punctuation is deleted rather than replaced, surface forms such as decimals and negative numbers may change during normalization. The LLM judges instead evaluate the generated answers directly without this string normalization.

EM compares sets of whitespace-delimited tokens, ignoring order and repetition, whereas precision, recall, and F1 use multiset token overlap. Sub-EM checks whether either normalized string is a substring of the other without requiring token boundaries. ROUGE-L is computed on normalized strings with \texttt{RougeScorer(metric=rougeL, use\_stemmer=True)}; stemming is used only for ROUGE-L. Consequently, explanatory answers can be semantically correct under the LLM judge while receiving lower strict lexical scores. Tables~\ref{tab:em_scores}--\ref{tab:rouge_scores} report EM, F1, and ROUGE-L, and Table~\ref{tab:llm_judge_qwen} provides the supplementary \texttt{Qwen3-30B-A3B-Instruct-2507} judge results.

\begin{table*}[h]
    \centering
    \resizebox{\textwidth}{!}{
        \small 
        \setlength{\tabcolsep}{3pt}
        \renewcommand{\arraystretch}{1.2}
        \begin{tabular}{l|cccccc|ccccccc|c}
        \toprule
        \multirow{2}{*}{\textbf{Method}} & \multicolumn{6}{c|}{\textbf{Test-Time Learning (TTL)}} & \multicolumn{7}{c|}{\textbf{Accurate Retrieval (AR)}} & \\
        & \textbf{Bank77} & \textbf{Clinic} & \textbf{NLU} & \textbf{Pub} & \textbf{T-C} & \textbf{T-F} & \textbf{Convo} & \textbf{Hotpot} & \textbf{LoCo} & \textbf{LME} & \textbf{MSC} & \textbf{PerL} & \textbf{SQuAD} & \textbf{Avg.} \\
        \midrule
        \multicolumn{15}{l}{\textit{Baselines}} \\
        Concat & 66.00 & 79.00 & 71.00 & 57.80 & 70.00 & 18.00 & 0.45 & 30.47 & 13.95 & 18.20 & 27.00 & 5.50 & 40.68 & 38.31 \\
        RAG ($k=20$) & 4.00 & 12.00 & 2.00 & 17.50 & 0.00 & 0.00 & 3.79 & 0.78 & 0.20 & 0.40 & 1.40 & 1.00 & 0.34 & 3.34 \\
        MemAgent & 24.00 & 13.00 & 23.00 & 11.70 & 61.00 & 37.00 & 2.90 & 14.06 & 2.92 & 8.80 & 5.20 & 1.00 & 3.31 & 15.99 \\
        MemAgent-woq & 27.00 & 41.00 & 34.00 & 0.00 & 69.00 & 42.00 & 0.00 & 0.00 & 0.05 & 0.00 & 0.40 & 0.00 & 0.17 & 16.43 \\
        Mem1 & 0.00 & 4.00 & 0.00 & 0.00 & 7.00 & 0.00 & 0.00 & 0.00 & 0.05 & 0.00 & 0.00 & 0.00 & 0.00 & 0.85 \\
        A-MEM & 71.00 & 77.00 & 71.00 & 53.50 & 32.00 & 13.00 & 13.39 & 44.53 & 9.42 & 15.60 & 24.00 & 9.50 & 34.62 & 36.04 \\
        Mem-$\alpha$ & 83.00 & 78.00 & 67.00 & 57.10 & 78.00 & 62.00 & 0.00 & 0.00 & 0.00 & 0.00 & 0.00 & 0.00 & 0.00 & 32.70 \\
        Mem-T & 58.00 & 35.00 & 46.00 & 21.10 & 0.00 & 3.00 & 15.40 & 41.41 & 15.61 & 17.80 & \textbf{33.00} & 11.50 & 39.79 & 25.97 \\
        \midrule
        \multicolumn{15}{l}{\textit{Ablations}} \\
        UMA (w/o RL \& Phase I) & 26.00 & 27.00 & 22.00 & 39.00 & 12.00 & 7.00 & 18.08 & 34.38 & 16.52 & 20.20 & 15.60 & 10.50 & 46.68 & 22.69 \\
        UMA (w/o Phase I) & 84.00 & 79.00 & 75.00 & 47.30 & 81.00 & 25.00 & 19.42 & 43.75 & 15.51 & 25.00 & 12.60 & 11.25 & 48.40 & 43.63 \\
        UMA (w/o RL) & 76.00 & 72.00 & 61.00 & 37.30 & 69.00 & 20.00 & 19.42 & 41.41 & 16.72 & 25.00 & 23.80 & 9.00 & 44.47 & 39.62 \\
        UMA (w/o Raw-Context) & 69.00 & 71.00 & 59.00 & 36.30 & 71.00 & 44.00 & 18.74 & 40.46 & 15.38 & 26.46 & 21.20 & 8.25 & \textbf{59.52} & 41.56 \\
        UMA (Global Group) & 69.00 & 77.00 & \textbf{82.00} & \textbf{69.10} & 87.00 & 47.00 & 19.20 & 40.62 & \textbf{19.54} & 14.80 & 28.40 & 6.50 & 30.73 & 45.45 \\
        UMA (Two-Stage) & 84.00 & 84.00 & 69.00 & 44.30 & 77.00 & 43.00 & 19.20 & 43.75 & 13.75 & 24.40 & 28.00 & 11.00 & 37.62 & 44.54 \\
        \midrule
        \textbf{UMA-Generalist (16k)} & \textbf{87.00} & \textbf{91.00} & \textbf{82.00} & 67.20 & \textbf{95.00} & \textbf{75.00} & \textbf{20.98} & \underline{\textbf{48.44}} & 17.62 & \textbf{30.60} & 29.20 & \textbf{12.25} & 56.32 & \textbf{54.82} \\
        \midrule
        \multicolumn{15}{l}{\textit{Session-budget scaling}} \\
        UMA-Generalist (8k) & 79.00 & 88.00 & 75.00 & 65.70 & 95.00 & 70.00 & 19.87 & 42.97 & 16.16 & 27.80 & 29.40 & \underline{13.25} & 55.60 & 52.13 \\
        UMA-Generalist (32k) & \underline{90.00} & \underline{92.00} & \underline{84.00} & \underline{68.60} & \underline{97.00} & \underline{83.00} & \underline{21.65} & 47.66 & \underline{18.23} & \underline{31.20} & \underline{31.20} & 12.25 & \underline{56.46} & \underline{56.40} \\
        \bottomrule
        \end{tabular}
    }
    \caption{Exact Match (EM) scores across all benchmarks. Unless otherwise specified, all methods use a 16k context budget. The UMA rows use UMA-Generalist. Bold numbers indicate the best result among methods using the 16k budget; underlined numbers indicate the best result among the UMA-Generalist 8k, 16k, and 32k variants.}
    \label{tab:em_scores}
\end{table*}

\begin{table*}[h]
    \centering
    \resizebox{\textwidth}{!}{
        \small 
        \setlength{\tabcolsep}{3pt}
        \renewcommand{\arraystretch}{1.2}
        \begin{tabular}{l|cccccc|ccccccc|c}
        \toprule
        \multirow{2}{*}{\textbf{Method}} & \multicolumn{6}{c|}{\textbf{Test-Time Learning (TTL)}} & \multicolumn{7}{c|}{\textbf{Accurate Retrieval (AR)}} & \\
        & \textbf{Bank77} & \textbf{Clinic} & \textbf{NLU} & \textbf{Pub} & \textbf{T-C} & \textbf{T-F} & \textbf{Convo} & \textbf{Hotpot} & \textbf{LoCo} & \textbf{LME} & \textbf{MSC} & \textbf{PerL} & \textbf{SQuAD} & \textbf{Avg.} \\
        \midrule
        \multicolumn{15}{l}{\textit{Baselines}} \\
        Concat & 70.00 & 79.00 & 71.00 & 57.80 & 70.00 & 18.00 & 8.65 & 49.27 & 25.89 & 27.84 & 48.21 & 18.00 & 53.57 & 45.94 \\
        RAG ($k=20$) & 25.12 & 30.82 & 20.90 & 18.92 & 3.63 & 2.04 & 38.66 & 15.57 & 16.90 & 17.62 & 23.33 & 13.92 & 5.29 & 17.90 \\
        MemAgent & 24.40 & 13.00 & 23.00 & 26.01 & 61.00 & 37.00 & 33.55 & 26.34 & 21.96 & 27.11 & 25.11 & 8.42 & 14.88 & 26.29 \\
        MemAgent-woq & 27.00 & 41.00 & 34.00 & 0.82 & 69.67 & 42.04 & 12.13 & 2.71 & 2.96 & 5.62 & 8.77 & 2.22 & 3.38 & 19.41 \\
        Mem1 & 0.00 & 4.00 & 0.00 & 0.44 & 7.00 & 0.00 & 14.22 & 1.66 & 3.37 & 5.60 & 5.69 & 1.44 & 1.48 & 3.45 \\
        A-MEM & 79.00 & 79.00 & 71.67 & 53.57 & 32.00 & 13.00 & 37.70 & 61.99 & 28.61 & 28.84 & 49.10 & 30.05 & 57.46 & 47.84 \\
        Mem-$\alpha$ & 83.00 & 78.00 & 67.02 & 57.10 & 78.00 & 62.00 & 16.18 & 8.51 & 5.50 & 9.33 & 9.58 & 4.37 & 5.18 & 37.21 \\
        Mem-T & 61.86 & 42.94 & 52.16 & 32.97 & 7.09 & 5.14 & 31.21 & 59.06 & 35.90 & 33.88 & \textbf{56.48} & \textbf{32.95} & 53.62 & 38.87 \\
        \midrule
        \multicolumn{15}{l}{\textit{Ablations}} \\
        UMA (w/o RL \& Phase I) & 26.00 & 27.50 & 22.00 & 40.20 & 12.00 & 7.00 & 40.94 & 52.73 & 36.70 & 28.82 & 28.36 & 26.01 & 62.08 & 31.56 \\
        UMA (w/o Phase I) & 84.67 & 79.25 & 75.00 & 47.35 & 81.00 & 25.00 & 44.21 & 59.53 & 34.57 & 36.66 & 21.08 & 28.22 & 63.18 & 52.28 \\
        UMA (w/o RL) & 78.12 & 72.00 & 61.67 & 39.64 & 69.00 & 20.00 & 43.38 & 56.81 & 32.96 & 36.32 & 46.73 & 24.01 & 60.38 & 49.31 \\
        UMA (w/o Raw-Context) & 69.00 & 71.00 & 59.00 & 36.30 & 71.00 & 44.00 & 43.72 & 56.64 & 30.78 & 33.86 & 40.20 & 29.39 & 59.52 & 49.57 \\
        UMA (Global Group) & 69.01 & 77.00 & \textbf{82.01} & \textbf{69.11} & 87.01 & 47.03 & 44.01 & 56.67 & \textbf{40.18} & 27.25 & 52.04 & 30.12 & 50.82 & 56.33 \\
        UMA (Two-Stage) & 85.07 & 84.67 & 70.17 & 45.47 & 77.00 & 43.00 & 44.36 & 58.73 & 33.03 & 37.53 & 49.98 & 28.86 & 49.78 & 54.43 \\
        \midrule
        \textbf{UMA-Generalist (16k)} & \textbf{87.00} & \textbf{91.00} & 82.00 & 67.20 & \textbf{95.00} & \textbf{75.00} & \textbf{46.33} & \underline{\textbf{65.42}} & 39.39 & \underline{\textbf{41.73}} & 52.59 & \underline{32.88} & \textbf{71.37} & \textbf{65.15} \\
        \midrule
        \multicolumn{15}{l}{\textit{Session-budget scaling}} \\
        UMA-Generalist (8k) & 79.00 & 88.00 & 75.00 & 65.70 & 95.00 & 70.00 & 44.16 & 56.51 & 37.05 & 37.34 & 51.03 & 32.61 & 70.64 & 61.69 \\
        UMA-Generalist (32k) & \underline{90.00} & \underline{92.00} & \underline{84.00} & \underline{68.60} & \underline{97.00} & \underline{83.00} & \underline{47.34} & 65.36 & \underline{40.08} & 41.11 & \underline{54.24} & 32.49 & \underline{71.54} & \underline{66.67} \\
        \bottomrule
        \end{tabular}
    }
    \caption{F1 scores across all benchmarks. Unless otherwise specified, all methods use a 16k context budget. The UMA rows use UMA-Generalist. Bold numbers indicate the best result among methods using the 16k budget; underlined numbers indicate the best result among the UMA-Generalist 8k, 16k, and 32k variants.}
    \label{tab:f1_scores}
\end{table*}

\begin{table*}[h]
    \centering
    \resizebox{\textwidth}{!}{
        \small 
        \setlength{\tabcolsep}{3pt}
        \renewcommand{\arraystretch}{1.2}
        \begin{tabular}{l|cccccc|ccccccc|c}
        \toprule
        \multirow{2}{*}{\textbf{Method}} & \multicolumn{6}{c|}{\textbf{Test-Time Learning (TTL)}} & \multicolumn{7}{c|}{\textbf{Accurate Retrieval (AR)}} & \\
        & \textbf{Bank77} & \textbf{Clinic} & \textbf{NLU} & \textbf{Pub} & \textbf{T-C} & \textbf{T-F} & \textbf{Convo} & \textbf{Hotpot} & \textbf{LoCo} & \textbf{LME} & \textbf{MSC} & \textbf{PerL} & \textbf{SQuAD} & \textbf{Avg.} \\
        \midrule
        \multicolumn{15}{l}{\textit{Baselines}} \\
        Concat & 70.00 & 79.00 & 71.00 & 57.80 & 70.00 & 18.00 & 8.63 & 50.44 & 25.97 & 27.76 & 48.79 & 18.36 & 54.00 & 46.13 \\
        RAG ($k=20$) & 25.13 & 30.82 & 20.90 & 18.92 & 3.63 & 2.04 & 36.55 & 15.47 & 16.55 & 16.30 & 23.17 & 13.92 & 5.40 & 17.60 \\
        MemAgent & 24.40 & 13.00 & 23.00 & 26.04 & 61.00 & 37.00 & 32.31 & 25.71 & 21.51 & 26.59 & 25.17 & 8.35 & 15.15 & 26.10 \\
        MemAgent-woq & 27.00 & 41.00 & 34.00 & 0.83 & 69.67 & 42.04 & 10.80 & 2.74 & 2.88 & 4.81 & 8.78 & 2.33 & 3.44 & 19.25 \\
        Mem1 & 0.00 & 4.00 & 0.00 & 0.45 & 7.00 & 0.00 & 11.63 & 1.77 & 3.28 & 4.70 & 5.36 & 1.54 & 1.54 & 3.17 \\
        A-MEM & 79.00 & 79.00 & 71.67 & 53.57 & 32.00 & 13.00 & 37.56 & 61.99 & 28.62 & 28.80 & 50.01 & 30.53 & 57.80 & 47.97 \\
        Mem-$\alpha$ & 83.00 & 78.00 & 67.02 & 57.10 & 78.00 & 62.00 & 14.33 & 8.49 & 5.26 & 8.43 & 9.71 & 4.29 & 5.22 & 36.99 \\
        Mem-T & 61.87 & 42.98 & 52.16 & 32.97 & 7.01 & 5.14 & 31.17 & 59.43 & 35.75 & 33.42 & \textbf{57.49} & \textbf{34.06} & 54.18 & 39.05 \\
        \midrule
        \multicolumn{15}{l}{\textit{Ablations}} \\
        UMA (w/o RL \& Phase I) & 26.00 & 27.50 & 22.00 & 40.20 & 12.00 & 7.00 & 40.86 & 52.78 & 36.34 & 28.76 & 29.18 & 26.24 & 63.03 & 31.68 \\
        UMA (w/o Phase I) & 84.67 & 79.25 & 75.00 & 47.35 & 81.00 & 25.00 & 44.16 & 59.34 & 34.15 & 36.51 & 21.29 & 28.57 & 64.20 & 52.34 \\
        UMA (w/o RL) & 78.12 & 72.00 & 61.67 & 39.64 & 69.00 & 20.00 & 43.28 & 56.79 & 32.87 & 36.17 & 48.05 & 24.14 & 61.38 & 49.47 \\
        UMA (w/o Raw-Context) & 69.00 & 71.00 & 59.00 & 36.30 & 71.00 & 44.00 & 43.42 & 56.44 & 31.68 & 33.80 & 41.20 & 28.19 & 59.25 & 49.56 \\
        UMA (Global Group) & 69.01 & 77.00 & \textbf{82.01} & \textbf{69.11} & 87.01 & 47.03 & 43.91 & 57.16 & \textbf{39.85} & 27.33 & 52.41 & 29.97 & 51.05 & 56.37 \\
        UMA (Two-Stage) & 85.07 & 84.67 & 70.17 & 45.47 & 77.00 & 43.00 & 44.40 & 59.47 & 32.72 & 37.40 & 50.89 & 29.36 & 50.59 & 54.63 \\
        \midrule
        \textbf{UMA-Generalist (16k)} & \textbf{87.00} & \textbf{91.00} & 82.00 & 67.20 & \textbf{95.00} & \textbf{75.00} & \textbf{46.39} & \underline{\textbf{66.16}} & 39.13 & \underline{\textbf{41.52}} & 53.28 & \underline{33.14} & \textbf{72.04} & \textbf{65.30} \\
        \midrule
        \multicolumn{15}{l}{\textit{Session-budget scaling}} \\
        UMA-Generalist (8k) & 79.00 & 88.00 & 75.00 & 65.70 & 95.00 & 70.00 & 44.30 & 57.25 & 36.79 & 37.20 & 51.90 & 32.88 & 71.28 & 61.87 \\
        UMA-Generalist (32k) & \underline{90.00} & \underline{92.00} & \underline{84.00} & \underline{68.60} & \underline{97.00} & \underline{83.00} & \underline{47.44} & 66.01 & \underline{39.85} & 40.81 & \underline{54.97} & 32.93 & \underline{72.34} & \underline{66.84} \\
        \bottomrule
        \end{tabular}
    }
    \caption{ROUGE-L scores across all benchmarks. Unless otherwise specified, all methods use a 16k context budget. The UMA rows use UMA-Generalist. Bold numbers indicate the best result among methods using the 16k budget; underlined numbers indicate the best result among the UMA-Generalist 8k, 16k, and 32k variants.}
    \label{tab:rouge_scores}
\end{table*}

\begin{table*}[h]
    \centering
    \resizebox{\textwidth}{!}{
        \small 
        \setlength{\tabcolsep}{3pt}
        \renewcommand{\arraystretch}{1.2}
        \begin{tabular}{l|cccccc|ccccccc|c}
        \toprule
        \multirow{2}{*}{\textbf{Method}} & \multicolumn{6}{c|}{\textbf{Test-Time Learning (TTL)}} & \multicolumn{7}{c|}{\textbf{Accurate Retrieval (AR)}} & \\
        & \textbf{Bank77} & \textbf{Clinic} & \textbf{NLU} & \textbf{Pub} & \textbf{T-C} & \textbf{T-F} & \textbf{Convo} & \textbf{Hotpot} & \textbf{LoCo} & \textbf{LME} & \textbf{MSC} & \textbf{PerL} & \textbf{SQuAD} & \textbf{Avg.} \\
        \midrule
        \multicolumn{15}{l}{\textit{Baselines}} \\
        Concat & 72.00 & 79.00 & 71.00 & 57.80 & 71.00 & 18.00 & 5.36 & 65.62 & 33.43 & 33.00 & 61.20 & 42.50 & 64.42 & 51.87 \\
        RAG ($k=20$) & 81.00 & 70.00 & 65.00 & 53.30 & 10.00 & 10.00 & 51.34 & 77.34 & 57.35 & \textbf{61.20} & 55.80 & \textbf{85.25} & 25.56 & 54.09 \\
        MemAgent & 26.00 & 14.00 & 24.00 & 51.20 & 61.00 & 37.00 & 60.27 & 51.56 & \textbf{64.15} & 59.40 & 53.00 & 71.50 & 78.06 & 50.09 \\
        MemAgent-woq & 27.00 & 41.00 & 34.00 & 47.00 & 70.00 & 44.00 & 2.68 & 14.84 & 29.71 & 9.80 & 29.60 & 26.75 & 25.80 & 30.94 \\
        Mem1 & 0.00 & 4.00 & 0.00 & 45.40 & 7.00 & 0.00 & 1.56 & 6.25 & 22.41 & 8.00 & 9.60 & 7.75 & 18.43 & 10.03 \\
        A-MEM & 83.00 & 80.00 & 72.00 & 53.70 & 32.00 & 15.00 & 53.35 & 72.22 & 50.25 & 51.61 & 64.21 & 80.00 & \textbf{89.46} & 61.29 \\
        Mem-$\alpha$ & 83.00 & 78.00 & 71.00 & 57.40 & 78.00 & 60.00 & 8.48 & \textbf{78.91} & 48.64 & 55.60 & 56.60 & 79.50 & 83.40 & 64.50 \\
        Mem-T & 69.00 & 53.00 & 59.00 & 54.90 & 23.00 & 12.00 & 51.34 & 69.53 & 45.52 & 55.00 & 73.00 & 67.75 & 67.76 & 53.91 \\
        \midrule
        \multicolumn{15}{l}{\textit{Ablations}} \\
        UMA (w/o RL \& Phase I) & 34.00 & 28.00 & 24.00 & 40.20 & 23.00 & 15.00 & 79.91 & 62.50 & 54.88 & 41.00 & 38.20 & 65.50 & 82.23 & 45.26 \\
        UMA (w/o Phase I) & 84.00 & 79.00 & 74.00 & 47.70 & 81.00 & 25.00 & 86.61 & 71.09 & 47.43 & 51.40 & 28.80 & 69.75 & 83.71 & 63.81 \\
        UMA (w/o RL) & 76.00 & 72.00 & 61.00 & 38.80 & 70.00 & 20.00 & 85.49 & 66.41 & 47.99 & 50.80 & 66.20 & 62.50 & 80.71 & 61.38 \\
        UMA (w/o Raw-Context) & 71.00 & 73.00 & 63.00 & 40.30 & 73.00 & 45.00 & 71.43 & 62.50 & 44.86 & 37.60 & 63.00 & 51.25 & 71.97 & 59.07 \\
        UMA (Global Group) & 72.00 & 79.00 & \textbf{82.00} & \textbf{74.00} & 89.00 & 55.00 & 83.71 & 76.56 & 61.98 & 44.00 & \textbf{75.40} & 83.00 & 80.92 & 73.58 \\
        UMA (Two-Stage) & 84.00 & 84.00 & 69.00 & 45.00 & 78.00 & 43.00 & 85.49 & 69.53 & 45.97 & 52.20 & 67.60 & 71.50 & 66.45 & 66.29 \\
        \midrule
        \textbf{UMA-Generalist (16k)} & \textbf{87.00} & \textbf{91.00} & \textbf{82.00} & 67.20 & \textbf{95.00} & \textbf{75.00} & \textbf{88.17} & \underline{77.34} & 50.10 & \underline{54.80} & 68.00 & \underline{82.50} & 87.32 & \textbf{77.34} \\
        \midrule
        \multicolumn{15}{l}{\textit{Session-budget scaling}} \\
        UMA-Generalist (8k) & 79.00 & 88.00 & 75.00 & 65.70 & 95.00 & 70.00 & 83.04 & 66.41 & 48.04 & 46.80 & 65.40 & 79.75 & 86.53 & 72.97 \\
        UMA-Generalist (32k) & \underline{90.00} & \underline{92.00} & \underline{84.00} & \underline{68.60} & \underline{97.00} & \underline{83.00} & \underline{90.18} & 74.22 & \underline{50.86} & 51.80 & \underline{69.60} & 82.00 & \underline{87.87} & \underline{78.55} \\
        \bottomrule
        \end{tabular}
    }
    \caption{LLM-as-a-Judge summary scores using \texttt{Qwen3-30B-A3B-Instruct-2507} as the evaluator across all benchmarks. Unless otherwise specified, all methods use a 16k context budget. The UMA rows use UMA-Generalist. Bold numbers indicate the best result among methods using the 16k budget; underlined numbers indicate the best result among the UMA-Generalist 8k, 16k, and 32k variants.}
    \label{tab:llm_judge_qwen}
\end{table*}

\section{Manual Error Analysis}
\label{app:error_analysis}

To better characterize the remaining failure modes, we manually inspect incorrect predictions together with the corresponding Memory Bank snapshots, retrieval outputs, and final answers. We assign each case to the earliest failing stage: memory formation, retrieval/access, or final evidence use.

\paragraph{Ledger-QA.}
On the $H=50$ split of Ledger-QA, which contains 105 queries, UMA-Specialist achieves 45.71\% LLM-as-a-Judge accuracy under the DeepSeek-V3.2 evaluator and makes 57 errors. Among them, 37 cases (64.9\%) are memory-formation errors, including incorrect writes, incomplete aggregation, or stale/conflicting state entries, while 20 cases (35.1\%) are QA-time reasoning errors. A typical memory error is an incorrect numeric update: two same-day Shopping expenses should be aggregated as $145.50 + 25.99 = 171.49$, but the memory bank stores $166.49$, causing later range queries to fail even when the relevant memory entry is accessed. This confirms that Ledger-QA primarily stresses durable state maintenance rather than local span retrieval.

\paragraph{LoCoMo.}
We further sample 50 UMA-Generalist cases on LoCoMo marked incorrect by the DeepSeek-V3.2 LLM-as-a-Judge evaluator to analyze failures in a dialogue-centric retrieval setting. We find 12 missing-memory cases (24\%), where the required fact is never written or is overwritten by an incomplete memory; 19 retrieval/access failures (38\%), where the relevant memory exists but is not selected; and 19 final evidence-use errors (38\%), where correct evidence is retrieved but ignored or polluted by noisy raw-context retrieval. These results suggest complementary bottlenecks: Ledger-QA is dominated by memory-write accuracy, whereas LoCoMo additionally requires stronger memory addressing and final evidence selection.

\section{Numerical Results on Ledger-QA}
\label{app:ledger_results}

Table~\ref{tab:ledger_detailed} reports per-horizon LLM-as-a-Judge and EM results. Average F1, ROUGE-L, and Sub-EM scores are summarized in Table~\ref{tab:ledger_checkpoint_summary}. UMA-Generalist is evaluated without Ledger-QA training, while UMA-Specialist uses the Ledger-QA training set. All averages are unweighted means across the nine session horizons.

\begin{table*}[p]
    \centering
    \scriptsize
    \setlength{\tabcolsep}{4pt}
    \renewcommand{\arraystretch}{0.92}
    \resizebox{0.99\textwidth}{!}{
    \begin{tabular}{l|ccccccccc|c}
    \toprule
    \multirow{2}{*}{\textbf{Method}} & \multicolumn{9}{c|}{\textbf{Session Horizon $H$}} & \\
     & \textbf{2} & \textbf{5} & \textbf{10} & \textbf{20} & \textbf{30} & \textbf{40} & \textbf{50} & \textbf{100} & \textbf{500} & \textbf{Avg.} \\
    \midrule
    \multicolumn{11}{l}{\textbf{LLM-as-a-Judge}} \\
    \multicolumn{11}{l}{\textit{Baselines}} \\
    Concat & 63.11 & 24.39 & 21.36 & 8.82 & 7.77 & 12.38 & 12.38 & 7.69 & 1.92 & 17.76 \\
    RAG & 80.58 & 39.02 & 21.36 & 17.65 & 13.59 & 15.24 & 10.48 & 8.65 & 5.77 & 23.59 \\
    MemAgent & 81.55 & 55.28 & 42.72 & 39.22 & 39.81 & 32.38 & 32.38 & 29.81 & 11.54 & 40.52 \\
    MemAgent-woq & 82.52 & 34.96 & 9.71 & 7.84 & 12.62 & 8.57 & 7.62 & 4.81 & 2.88 & 19.06 \\
    Mem1 & 27.18 & 9.76 & 4.85 & 3.92 & 9.71 & 1.90 & 1.90 & 6.73 & 9.62 & 8.40 \\
    A-MEM & 33.01 & 24.39 & 19.42 & 11.76 & 10.68 & 8.57 & 8.57 & 0.96 & 1.92 & 13.25 \\
    Mem-$\alpha$ & 60.19 & 37.40 & 34.95 & 21.57 & 18.45 & 16.19 & 22.86 & 14.42 & 9.62 & 26.18 \\
    Mem-T & 58.25 & 39.84 & 25.24 & 13.73 & 10.68 & 8.57 & 13.33 & 11.54 & 2.88 & 20.45 \\
    \addlinespace[1pt]
    \multicolumn{11}{l}{\textit{Ablations}} \\
    UMA (w/o RL \& Phase I) & 18.45 & 11.38 & 8.74 & 8.82 & 7.77 & 8.57 & 5.71 & 24.04 & 14.42 & 11.99 \\
    UMA (w/o RL) & 58.25 & 37.40 & 29.13 & 24.51 & 31.07 & 26.67 & 32.38 & 25.00 & 9.62 & 30.45 \\
    UMA-Specialist (w/o Raw) & 83.50 & 61.79 & 59.22 & \textbf{60.78} & 52.43 & \textbf{46.67} & \textbf{48.57} & \textbf{34.62} & \textbf{25.00} & 52.51 \\
    \addlinespace[1pt]
    \multicolumn{11}{l}{\textit{Transfer without Ledger-QA training}} \\
    UMA-Generalist (16k) & 73.79 & 52.85 & 42.72 & 30.39 & 36.89 & 29.52 & 44.76 & 21.15 & 11.54 & 38.18 \\
    \addlinespace[1pt]
    \multicolumn{11}{l}{\textit{UMA-Specialist session-budget scaling}} \\
    UMA-Specialist (8k) & 83.50 & 59.35 & 54.37 & 50.00 & 39.81 & 40.00 & 33.33 & 22.12 & 20.19 & 44.74 \\
    \textbf{UMA-Specialist (16k)} & \underline{\textbf{87.38}} & \textbf{63.41} & \underline{\textbf{63.11}} & \underline{57.84} & \underline{\textbf{53.40}} & \textbf{46.67} & 45.71 & \textbf{34.62} & \textbf{25.00} & \textbf{53.02} \\
    UMA-Specialist (32k) & 85.44 & \underline{67.48} & 61.17 & 54.90 & 48.54 & \underline{50.48} & \underline{47.62} & \underline{38.46} & \underline{25.96} & \underline{53.34} \\
    \midrule
    \multicolumn{11}{l}{\textbf{Exact Match (EM)}} \\
    \multicolumn{11}{l}{\textit{Baselines}} \\
    Concat & 22.33 & 11.38 & 11.65 & 4.90 & 1.94 & 4.76 & 8.57 & 4.81 & 0.96 & 7.92 \\
    RAG & 0.97 & 0.00 & 0.00 & 0.00 & 0.00 & 0.00 & 0.00 & 0.00 & 0.00 & 0.11 \\
    MemAgent & 47.57 & 28.46 & 20.39 & 14.71 & 14.56 & 14.29 & 10.48 & 14.42 & 8.65 & 19.28 \\
    MemAgent-woq & 0.97 & 1.63 & 1.94 & 0.98 & 0.00 & 0.00 & 0.00 & 0.00 & 0.00 & 0.61 \\
    Mem1 & 0.97 & 3.25 & 0.97 & 2.94 & 5.83 & 0.95 & 0.00 & 0.00 & 6.73 & 2.40 \\
    A-MEM & 8.74 & 8.94 & 7.77 & 4.90 & 6.80 & 3.81 & 2.86 & 0.00 & 0.96 & 4.97 \\
    Mem-$\alpha$ & 0.97 & 0.00 & 0.00 & 0.00 & 0.00 & 0.00 & 0.00 & 0.00 & 0.00 & 0.11 \\
    Mem-T & 27.18 & 17.07 & 8.74 & 4.90 & 2.91 & 1.90 & 11.43 & 3.85 & 0.00 & 8.67 \\
    \addlinespace[1pt]
    \multicolumn{11}{l}{\textit{Ablations}} \\
    UMA (w/o RL \& Phase I) & 0.00 & 0.00 & 0.00 & 0.00 & 0.00 & 0.00 & 0.00 & 2.88 & 4.81 & 0.85 \\
    UMA (w/o RL) & 33.98 & 26.02 & 19.42 & 15.69 & 21.36 & 18.10 & 25.71 & 8.65 & 2.88 & 19.09 \\
    UMA-Specialist (w/o Raw) & \textbf{79.61} & 60.98 & 58.25 & \textbf{60.78} & 52.43 & \textbf{46.67} & \textbf{46.67} & \textbf{34.62} & 8.65 & \textbf{49.85} \\
    \addlinespace[1pt]
    \multicolumn{11}{l}{\textit{Transfer without Ledger-QA training}} \\
    UMA-Generalist (16k) & 64.08 & 49.59 & 40.78 & 28.43 & 33.01 & 27.62 & 37.14 & 19.23 & \textbf{9.62} & 34.39 \\
    \addlinespace[1pt]
    \multicolumn{11}{l}{\textit{UMA-Specialist session-budget scaling}} \\
    UMA-Specialist (8k) & 77.67 & 57.72 & 51.46 & 48.04 & 36.89 & 38.10 & 31.43 & 22.12 & \underline{13.46} & 41.88 \\
    \textbf{UMA-Specialist (16k)} & 77.67 & \textbf{62.60} & \underline{\textbf{62.14}} & \underline{57.84} & \underline{\textbf{53.40}} & \textbf{46.67} & 43.81 & 32.69 & 7.69 & 49.39 \\
    UMA-Specialist (32k) & \underline{79.61} & \underline{65.85} & 60.19 & 54.90 & 48.54 & \underline{50.48} & \underline{47.62} & \underline{38.46} & 11.54 & \underline{50.80} \\
    \bottomrule
    \end{tabular}
    }
    \caption{Per-horizon Ledger-QA results, grouped by evaluation metric for direct comparison across methods. LLM-as-a-Judge uses DeepSeek-V3.2. Bold numbers indicate the best result among methods using the 16k budget; underlined numbers indicate the best result among the UMA-Specialist 8k, 16k, and 32k variants.}
    \label{tab:ledger_detailed}
\end{table*}

\end{document}